\documentclass[letterpaper]{article}
\usepackage{aaai}
\usepackage{times}  
\usepackage{helvet}  
\usepackage{courier}  
\usepackage[hyphens]{url}  
\usepackage{graphicx} 
\usepackage{natbib}  
\usepackage{caption} 
\usepackage{amsthm}

\theoremstyle{plain}

\theoremstyle{definition}

\theoremstyle{remark}

\usepackage{algorithm}
\usepackage{algorithmic}

\usepackage{pgfplots}
\pgfplotsset{compat=1.18} 
\usepackage{multirow}
\usepackage{colortbl}
\usepackage{color}
\usepackage{subcaption} 
\usepackage{booktabs} 
\usepackage[capitalize]{cleveref}
\crefname{figure}{Figure}{Figure}
\usepackage{arydshln}
\usepackage{tikz}
\usepackage{tabularx}
\usepackage{longtable}
\usepackage{amsfonts}

\title{Stabilizing Long-term Multi-turn Reinforcement Learning with Gated Rewards}
\author{
    Zetian Sun\equalcontrib, 
    Dongfang Li\equalcontrib, 
    Zhuoen Chen, 
    Yuhuai Qin, 
    Baotian Hu
}
\affiliations{
    Harbin Institute of Technology (Shenzhen), Shenzhen, China
 \\
    zetiansun.cs@gmail.com, \\
    \{lidongfang, hubaotian\}@hit.edu.cn
}

\begin{document}
 
%

\maketitle
\begin{abstract}
\begin{quote}
Reward sparsity in long-horizon reinforcement learning (RL) tasks remains a significant challenge, while existing outcome-based reward shaping struggles to define meaningful immediate rewards without introducing bias or requiring explicit task decomposition. Alternatively, verification-based reward shaping uses stepwise critics, but misalignment between immediate rewards and long-term objectives can lead to reward hacking and suboptimal policies. In this work, we address this problem in the context of software engineering (SWE) tasks, where multi-turn reasoning and rule-based verification are critical. We introduce the SWE-oriented RL Framework, a unified system supporting multi-turn interaction, docker-based execution, and customizable reward functions. Additionally, we propose Gated Reward Accumulation (G-RA), a novel method that accumulates immediate rewards only when high-level (long-term) rewards meet a predefined threshold, ensuring stable RL optimization. Experiments on SWE-bench Verified and kBench demonstrate that G-RA leads to an increase in completion rates (47.6\% → 93.8\% and 22.0\% → 86.0\%) and modification rates (19.6\% → 23.8\% and 12.0\% → 42.0\%), while avoiding policy degradation caused by reward misalignment. Our findings highlight the importance of balanced reward accumulation in long-horizon RL and provide a practical solution.

\end{quote}
\end{abstract}
\section{Introduction}
\begin{figure}[t]
    \centering
    \includegraphics[scale=0.5]{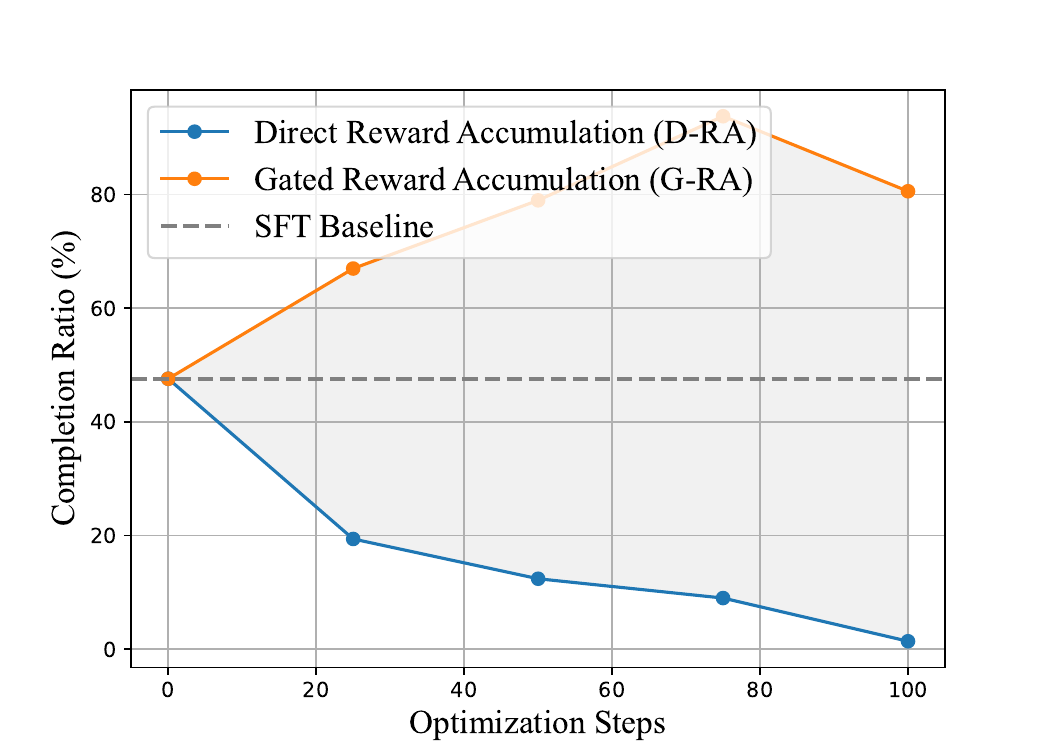}
    \vspace{-5pt}
    \caption{The completion ratio curve between D-RA and G-RA across different steps in SWE-bench Verified. We observe an opposite trend for the result of both methods: model trained with D-RA encounters model collapse, while model trained with G-RA achieves improved completion ratio.}
    \label{fig:cr_ra_curve}
    \vspace{-4mm}
\end{figure}

Reward sparsity remains a significant problem with respect to the optimization of long-horizon reinforcement learning~(RL) tasks. To acquire dense reward signal, previous works focus on outcome-based reward shaping methods, such as potential-based reward shaping and Hierarchical Reinforcement Learning~(HER). Specifically, potential-based reward shaping method defines the distance between current state and the final outcome to continualize the discrete sparse reward~\citep{DBLP:conf/icml/NgHR99,DBLP:conf/nips/WangYDSLU23}, HER method performs task decomposition and setting sub-goals towards the final outcome~\citep{DBLP:conf/iros/RanchodRK15,DBLP:conf/iclr/SharmaSRK19,DBLP:conf/icml/ZhouZPLK24} to improve reward granularity. However, both methods require a direct link between intermediate reasoning steps and the final objective. This connection can also facilitate the design of the immediate rewards, aligning its optimization goal with the long-term objective, which is difficult when the relationship between intermediate reasoning steps and the final objective is not explicitly known or easily definable.

An alternative to outcome-based reward shaping is the verification-based reward shaping methods, where immediate rewards are derived from rule-based stepwise critics~(i.e., verifiable immediate rewards) that depend on individual actions rather than the long-term objective. These rewards are easy to define but heavily rely on human priors. For instance, Deepseek-R1~\citep{DBLP:journals/corr/abs-2501-12948} introduces an action format reward to encourage the language model policy to generate responses in a reason-acting format~\citep{DBLP:conf/iclr/YaoZYDSN023}, which has shown effective for reasoning tasks. In single-turn scenarios, verification-based reward shaping methods take the form of auxiliary outcome rewards and computes only after the reasoning process completes, which is similar to outcome-based rewards~\citep{DBLP:journals/corr/abs-2402-03300,DBLP:journals/corr/abs-2504-20571}. Here, the final reward is a linear combination of the outcome reward and auxiliary rewards, with the optimization goal framed as achieving Pareto optimality between them. However, in multi-turn scenarios, aligning verifiable immediate rewards with the long-term objective becomes challenging. While stepwise critics are individually verifiable, their misalignment with the final outcome reward can lead to severe reward over-optimization and shortcut learning~\citep{DBLP:journals/corr/abs-2503-15478,DBLP:journals/corr/abs-2504-20073}, particularly when obtaining positive rewards from stepwise critics is significantly easier than from the outcome reward, which phenomenon is further observed by our experiments and the shown in Figure~\ref{fig:cr_ra_curve}. This motivates us to study the following research question: \textbf{how to accumulate immediate rewards and long-term objectives and stabilize RL optimization?}

In this paper, we try to answer the question from the systematic and methodological perspective.  Specifically, we focus on software engineering (SWE) scenario, where the problem-solving process requires multi-turn interaction between the language model~(LM) policy and the task-specific repository, and the outcome correctness is judged in a rule-based way via execution-based verification. For the first time, we propose the \textbf{SWE-oriented RL Framework}, a unified framework that incorporates the docker-based environment that supports multi-turn policy interaction, and a RL training system that supports multi-turn RL optimization. Specifically, the interaction framework is designed as a four-level architecture. Once the LM policy generates a text-based action to interact with the environment, the action will be parsed and converted into the standard commands in the scaffold layer, then the command will be sent to the low-level interface layer, where the atomic executable actions are defined and implemented. The command will finally be executed in the lowest layer, i.e., the environment layer, where the environment status will be modified and the environmental feedback will be sent back to the upper layers. We design the environment to be more suitable for RL research, including hindsight principle that dangerous commands will not be prevented but dealt with afterwards, and high customizability for different settings including environmental feedback, reward functions, scaffolds and SWE-related tasks. We further propose the \textbf{G}ated \textbf{R}eward \textbf{A}ccumulation~(\textbf{G-RA}) method, a multi-level reward accumulation approach that the calculation of low-level rewards will be masked when the high-level rewards are not above the preset threshold. Specifically, we set the long-term objective to be the high-level reward, the verifiable immediate rewards to be low-level rewards, and perform the reward accumulation in a heuristic approach: if the patch is empty or not generated, the long-term reward will be negative, and the rewards derived from stepwise critics will be zero. 

We conduct experiments using the SWE-oriented RL Framework on benchmarks including SWE-bench Verified~\citep{DBLP:conf/iclr/JimenezYWYPPN24} and kBench~\citep{DBLP:conf/nips/MathaiHMNIYR24}, aiming at demonstrating the challenge of reward accumulation in long-horizon scenarios, and the effectiveness of our method, especially when the long-term objective is sparse and difficult. The results show that model trained with direct reward accumulation method encounters catastrophic policy degradation during RL training, which we attribute to the consequences of reward misalignment and reward hacking through further analysis. Compared with the supervised-finetuned baseline, model trained with G-RA achieves an increase in task completion rate~(CR) and repository modification rate~(MR) in SWE-bench Verified~($47.2\%\rightarrow93.8\%$ for CR, $19.6\%\rightarrow23.8\%$ for MR) and kBench-50~($22.0\%\rightarrow86.0\%$ for CR, $12.0\%\rightarrow42.0\%$ for MR), indicating that G-RA can effectively stable the RL optimization in long-horizon scenarios. Our contributions are as follows:
\begin{itemize}
    \item We propose the SWE-oriented RL Framework, a unified framework that supports SWE-related environment construction, multi-turn RL training and evaluation.
    \item We propose G-RA, a novel reward accumulation method that balance verifiable immediate rewards and long-term objectives to stabilize RL optimization.
    \item We explore the reward misalignment problem, underscoring the importance of carefully designing fine-grained verifiable rewards in long-horizon multi-turn reinforcement learning scenarios.
\end{itemize}

\section{Related Work}
\subsection{Multi-turn Interaction Framework}
Several multi-turn interaction framework has been proposed for tasks in different scenarios. OpenAI Gym~\citep{brockman2016openai} is a widely used platform that provides a standard API to communicate between RL algorithms and game-based environments, as well as a standard set of environments compliant with that API. Recently, multi-agent systems focus on the interaction between different agents, where an isolated environment that provides rule-based feedback is not needed~\citep{DBLP:conf/nips/LiHIKG23,DBLP:conf/icml/Du00TM24,DBLP:conf/acl/WangWSTS24,DBLP:conf/iclr/YaoSRN25}. There are also interaction frameworks that focus on agent-environment interaction. OpenHands is an open platform for tasks that requires agent-environment interaction in similar ways to those of software developer, including software engineering~\citep{DBLP:conf/iclr/JimenezYWYPPN24} and web browsing~\citep{DBLP:conf/iclr/ZhouX0ZLSCOBF0N24}. OS World~\citep{DBLP:conf/nips/XieZCLZCHCSLLXZ24} is a real computer environment for multimodal agents across various operating systems and different real-world computer use cases, focusing on GUI-grounded operations and supporting execution-based evaluation. Our work builds on these advances, aiming at incorporating RL-based optimization with multi-turn interaction between models and environment across different long-horizon SWE-related tasks.

\subsection{Reward Shaping for Long-Horizon Scenarios}
Reward shaping is useful for tackling long-horizon tasks where the agent needs to navigate complex sequences of actions to achieve a goal, which can help improve sample efficiency and result in faster convergence during RL optimization~\citep{DBLP:conf/nips/HuWJWCH0F20,DBLP:journals/access/IbrahimMJSO24}. Potential-base reward shaping defines a potential function to measure the distance between current progress towards the ultimate goal~\citep{DBLP:conf/nips/WangYDSLU23,DBLP:conf/aaai/Forbes024}. Hierarchical Reinforcement Learning~(HER) decomposes the long-horizon RL task into several subtasks, where each subtask is to be handled by several actions~\citep{DBLP:journals/jair/Dietterich00,DBLP:reference/ml/Hengst10,DBLP:conf/aaai/Nayyar025}. The rewards are then defined in task-level and action-level, as the LM policy is required to choose the optimal subtasks during the problem-solving process. Auxiliary rewards methods focus on integrate auxiliary rewards that reflect the designer's heuristics and domain knowledge with the long-term ultimate objective~\citep{gupta2023behavior}, which is more straight-forward and easier to operate. Our work focuses on reward design and reward shaping in complex long-horizon scenarios, and optimizes the auxiliary reward methods by constructing the multi-level reward accumulation approach.
\section{Preliminaries}
\subsection{Markov Decision Process~(MDP)}
Generally, an MDP is defined as a tuple~$\mathcal{M}:=\langle\mathcal{S},\mathcal{A},\mathcal{T},\mathcal{R},\mu\rangle$, where $\mathcal{S}$ is the state set, $\mathcal{A}$ is the action set, $\mathcal{T}:\mathcal{S}\times\mathcal{A}\rightarrow\mathcal{M}(\mathcal{S})$ is the transition probability function, $\mathcal{M}(\mathcal{S})$ is the set of distribution over $\mathcal{S}$, $\mathcal{R}:\mathcal{S}\times\mathcal{A}\rightarrow\mathcal{M}([\mathcal{R}_{\rm min},\mathcal{R}_{\rm max}])$ is the reward distribution function, $\mathcal{M}([\mathcal{R}_{\rm min},\mathcal{R}_{\rm max}])$ is the set of distributions supported on $[\mathcal{R}_{\rm min},\mathcal{R}_{\rm max}]$ and $\mu\in\mathcal{M}(\mathcal{S})$ is the initial state distribution. The LM policy is defined by a mapping $\pi:\mathcal{S}\rightarrow\mathcal{M}(\mathcal{A})$. At each timestep $t$, $\pi$ generate the next action $a_t$ given current state $s_t$, after which a stepwise reward $r_t=\mathcal{R}(s_t,a_t)$ is received, and the policy transitions to the next state $s_{t+1}$. This process produces a finite trajectory $\tau:=(s_0,a_0,r_0,s_1,a_1,r_1,...,s_T,a_T,r_T)$ throughout the multi-turn interaction. The discount factor $\gamma\in[0,1]$ is used to calculate the expected return of each trajectory.

\subsection{Group Relative Policy Optimization~(GRPO)} 
Reinforcement Learning~(RL) enables LM policies to learn through interaction and reward signals. The general RL optimization objective is formulated as
\begin{equation}
    J(\theta)=\mathbb{E}_{s\sim\mathcal{D},a\sim\pi_\theta(\cdot|s)}[\mathcal{R}(s,a)],
\end{equation}
where $\pi_\theta$ is the current LM policy, $s$ is the current state, $a$ is the sampled action, and $\mathcal{R}(s,a)$ is the reward function. Given the input prompt $q$, GRPO~\citep{DBLP:journals/corr/abs-2402-03300} perform RL optimization by sampling $G$ outputs $\{o_i\}$, and optimizes
\begin{equation}
    J_{GRPO}(\theta)=\mathbb{E}_{q,\{o_i\}}[J_{group}(\theta)],
\end{equation}
where
\begin{equation}
    J_{group}(\theta)=\frac{1}{G}\sum_{i=1}^G\min(\rho_iA_i,\hat{\rho_i}A_i)-\beta D_{KL},
\end{equation}
the importance sampling ratio $\rho_i=\frac{\pi_\theta(o_i|q)}{\pi_{\theta_{old}}(o_i|q)}$, the clipped ratio $\hat{\rho_i}={\rm clip}(\rho_i,1-\varepsilon,1+\varepsilon)$, $D_{KL}$ is the Kullback-Leibler divergence between policy $\pi_\theta$ and reference $\pi_{ref}$.

The advantage function $A_i$ of GRPO is calculated in a critic-free manner that is based on relative rewards of the outputs inside each group only, which can be formulated as
\begin{equation}
    A_i=\frac{r_i-{\rm mean}(\{r_j\})}{{\rm std}(\{r_j\})},
\end{equation}
given $r_i$ as the reward of the $i_{th}$ output inside the group, $\{r_j\}$ as the rewards of each output inside the group.
\section{SWE-Oriented RL Framework}\label{sec:environment}
\begin{figure*}[t]
    \centering
    \includegraphics[width=\textwidth]{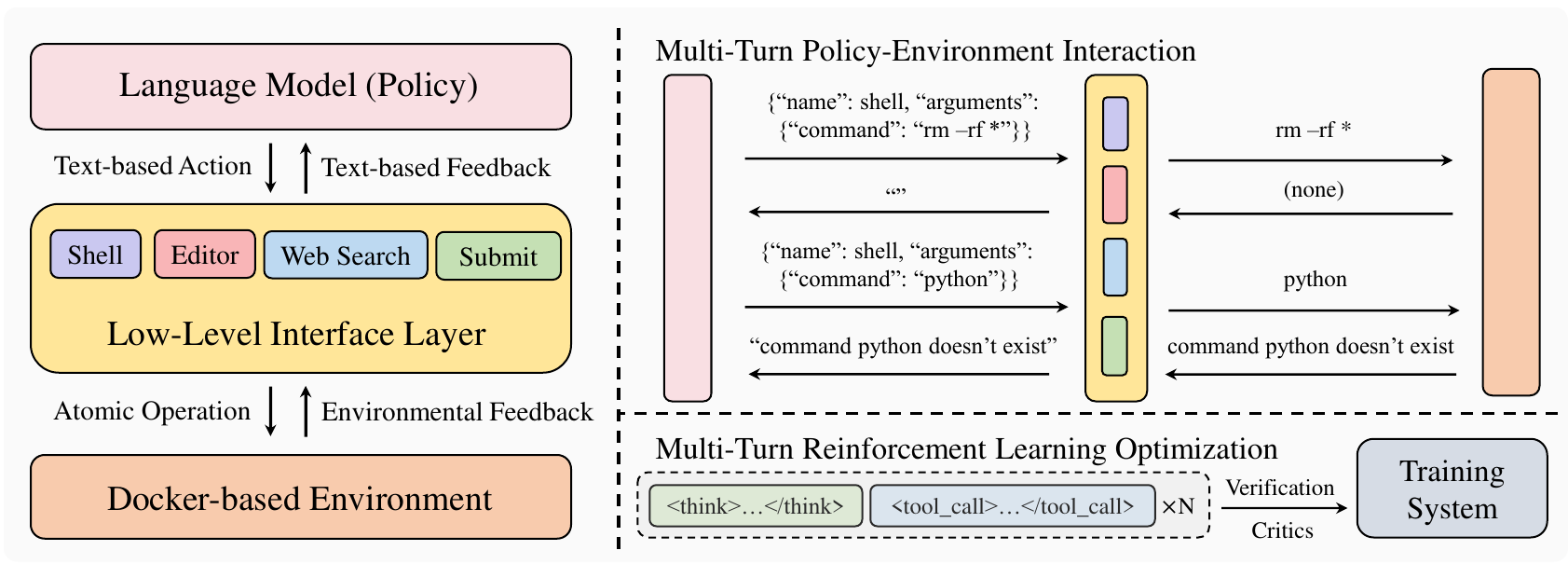}
    \vspace{-5pt}
    \caption{The illustration of our SWE-oriented RL Framework. Once the LM policy generates a text-based action, it will be executed in the environment by being captured by the scaffolds first, then being decomposed into atomic operations in the scaffold layer~(where defines and implements the different scaffolds) and low-level interface layer~(where defines and implements the atomic operations), respectively. Through multi-turn interaction between LM policy and the environment, the trajectories will be used for reward labeling~(including long-term objective and verifiable immediate rewards) and RL training. }
    \label{fig:environment}
\end{figure*}

In the following section, we present the detailed design of the RL framework for model optimization with respect to long-horizon multi-turn reinforcement learning. The framework is designed for tasks in SWE scenario and the illustration of the environment is shown in Figure~\ref{fig:environment}.
\subsection{Task Definition}
 We describe the interaction and problem-solving process as an MDP. Given a task-specific code repository $\mathcal{CR}$, the state $s_t\in\mathcal{S}$ denotes the status of the environment at timestep $t$, including the policy-environment interaction history, the status of task-specific repository, the installed python packages and else. The action $a_t=\pi(\cdot|s_t)\in\mathcal{A}$ is the output of the current policy $\pi$ given current state $s_t$. Specifically, we construct the context as the combination of previous actions $[a_1, a_2,...,a_{t-1}]$ and environmental feedbacks following a pre-defined multi-turn chat template, then ask the LM policy to generate the text-based action. The state transition $\mathcal{T}$ requires interactions between the LM policy and the isolated environment. For each action $a$, we transform the text-based action into an executable action by requesting a specific instance from a pre-defined set of scaffolds, which is then transformed into atomic operations. Then, the operations are executed in the environment, after which an environmental feedback is received as the action response. The outcome reward $\mathcal{R}$ is execution-based. There are also verification-based rewards like the correctness of action format and the correctness of tool usage for each intermediate state, which are useful to provide fine-grained and meaningful feedback.

\subsection{Overall Framework}\label{sec:env:scaffolds}
The interaction framework is a four-level architecture, including the policy layer, the scaffold layer, the low-level interface layer and the environment layer. The policy layer receives text-based feedback from lower layers, and sends text-based action to lower layers. The scaffold layer transforms the policy-generated text-based actions into standard executable operations by utilizing one of the pre-defined scaffolds with specific arguments. Then, the low-level interface layer transforms the standard executable operations into a sequence of atomic operations, for example, reading or writing files and executing shell commands. The atomic operations received from the low-level interface layer are executed in the environment layer, where consists of the isolated Linux docker container and the task-specific code repository. Finally, the raw feedback from the environment layer is wrapped and submitted to the upper layers.

The definition of scaffolds are customizable, serving different downstream scenarios and research purposes. In this paper, we define four basic scaffolds that allows LM policy to interact with environment based on its parametric knowledge and information observed from environment or external sources. We introduce the scaffolds as follows, more details are shown in Appendix~\ref{sec:app:scaffolds}.

\paragraph{Shell.} We provide the Bash shell scaffold for the LM policy to interact with the environment via command execution. Shell is the most commonly-used scaffold for SWE-oriented environment, as applied in Openhands~\citep{DBLP:conf/iclr/0001LSXTZPSLSTL25} and SWE-agent~\citep{DBLP:conf/nips/YangJWLYNP24}. 

\paragraph{Editor.} We provide the Editor scaffold for viewing, editing and creating file with specific contents, as an alternative to the ineffective built-in commands derived from bash shell. 

\paragraph{Web Search.} The Web Search scaffold is designed to support real-time information updates and active retrieval of external information apart from the isolated environment. 

\paragraph{Submit.} When the problem has been handled, LM policy is asked to request the Submit scaffold to generate git patch and finish the task. The Submit scaffold can only be successfully requested once, and the running environment will be immediately terminated after the git patch being generated.

\subsection{Additional Features}
Our framework incorporates several key design choices to enhance robustness, flexibility, and scalability in RL-driven problem solving. Specifically, we introduce four core features: (1) a hindsight principle for failure handling, (2) customizable scaffolds for task adaptation, (3) scalable scenarios via automated environment orchestration, and (4) parallel generation for efficient multi-instance training. We provide more details in Appendix~\ref{sec:app:environment}.
\begin{table*}[htbp]
\small
\centering
\begin{tabular}{lcccccc}
\toprule
\multirow{2}{*}{Method} & \multicolumn{3}{c}{SWE-bench Verified}    & \multicolumn{3}{c}{kBench-50} \\
            & CR (\%,$\uparrow$) & MR (\%,$\uparrow$) & RR (\%,$\uparrow$) & CR (\%,$\uparrow$) & MR (\%,$\uparrow$) & RR (\%,$\uparrow$) \\ 
\rowcolor{gray!12}
\multicolumn{7}{c}{\textbf{Proprietary Models}} \\
Deepseek-V3 & 75.2 & 62.6 & 16.4 & 90.0 & 88.0 & 8.0 \\
Deepseek-R1 & 18.8 & 18.8 & 5.4 & 98.0 & 88.0 & 8.0 \\
GPT-4o & 54.6 & 42.0 & 7.6 & 92.0 & 78.0 & 8.0\\
Gemini-2.5-Pro & 19.6 & 10.6 & 3.6 & 52.0  & 44.0  & 8.0  \\

\rowcolor{gray!12}
\multicolumn{7}{c}{\textbf{Open-Source Models}} \\
SFT & 47.6 & 19.6 & 0.2 & 22.0 & 12.0 & 0.0 \\
\midrule
D-RA~($25$ steps)  & 19.4 & 6.6 & 0.0 & 16.0 & 12.0 & 0.0         \\
D-RA~($50$ steps)  & 12.4 & 4.6 & 0.0 & 10.0 & 8.0 & 0.0        \\
D-RA~($75$ steps)  & 9.0 & 4.0 & 0.0 & 12.0 & 10.0 & 0.0         \\
D-RA~($100$ steps) & 1.4 & 0.2 & 0.0 & 0.0 & 0.0 & 0.0          \\
\midrule
G-RA~($25$ steps)  & 67.0 & 23.8 & 0.2 & 36.0 & 20.0 & 0.0       \\
G-RA~($50$ steps)  & 79.0 & 22.2 & 0.2 & 48.0 & 20.0 & 0.0        \\
G-RA~($75$ steps)  & 93.8 & 22.4 & 0.2 & 86.0 & 42.0 & 0.0        \\
G-RA~($100$ steps) & 80.6 & 17.2 & 0.2 & 84.0 & 32.0 & 0.0        \\
\bottomrule
\end{tabular}
\caption{The results on SWE-bench Verified and kBench-50 with proprietary models and open-source models. We report the Completion Rate~(CR, \%), Modification Rate~(MR, \%) and Resolution Rate~(RR, \%) for both benchmarks. For open-source models, we report the evaluation results of models trained by RL every $25$ optimization steps under different reward accumulation strategies (D-RA and G-RA). The results are averaged across three random seeds.
}
\label{tab:experiment-results}
\vspace{-4mm}
\end{table*}

\section{Gated Reward Accumulation~(G-RA)}
 In this work, we propose the Gated Reward Accumulation~(G-RA), an accumulation method to improve reward design and reward accumulation between the long-term objective and the verifiable immediate rewards in multi-turn long-horizon scenario. When the reward of high priority is of negative value, the rewards of lower priority will be masked and neglected. When the reward of high priority is of positive value, the rewards of lower priority will be normally calculated and be integrated to the final reward calculation. 
 
 Formally, given state $s\in\mathcal{S}$, action $a\in\mathcal{A}$, let $\{\mathcal{R}^{(1)},\mathcal{R}^{(2)},...,\mathcal{R}^{(n)}\}\in\mathcal{R}$ be a set of $n$ different reward functions, including the outcome reward and the rule-based stepwise critics. For each reward, we define the reward priority $\{o^{(1)},o^{(2)},...,o^{(n)}\}\in\mathbb{N}^+$, which represents that the reward priority of reward $\mathcal{R}^{(i)}$ is $o^{(i)}$. We define the gated value $\{gv^{(1)},gv^{(2)},...,gv^{(n)}\}\in\mathbb{R}$, which represents that given reward function $\mathcal{R}^{(i)}\in\mathcal{R}$, reward $\mathcal{R}^{(i)}(s,a)$ will be treat as a positive value if $\mathcal{R}^{(i)}(s,a)\geq gv^{(i)}$, otherwise the reward will be treated as a negative value. Given a pair of rewards $\{\mathcal{R}^{(i)},\mathcal{R}^{(j)}\}\in\mathcal{R}$ that satisfies $o^{(i)}<o^{(j)}$, the Gated Reward Accumulation method can be formulated as:
\begin{equation}
    \mathcal{R}^{(i)}(s,a)=\left\{
    \begin{array}{rcl}
        &0 &\mathcal{R}^{(j)}(s,a)\leq gv^{(j)}\\
        &\mathcal{R}^{(i)}(s,a)& \rm otherwise
    \end{array},
    \right.
\end{equation}

 Specifically, we introduce G-RA in SWE scenario. First of all, we define the reward functions as follows.
 \paragraph{Outcome Reward $\mathcal{R}^{(1)}$~(Long-Term Objective).} The outcome reward function is defined as whether the generated patch can successfully handle the problem for the given task repository, which is verified by the evaluation script used in~\cite{DBLP:conf/iclr/JimenezYWYPPN24}. We follow the official evaluation method to evaluate the correctness of the generated patch. Given limited interaction times~(maximum turn number), the outcome reward is $-2$ if the LM policy does not successfully call the Submit scaffold; $-1$ if the LM policy generates an empty patch, i.e., the LM policy does not modify the task repository before calling the Submit scaffold; $0$ if the generated patch fails to pass the test; $10$ if the generated patch works and passes the test. We set $o^{(1)}=3$ and $gv^{(1)}=0$.

 \paragraph{Action Format Reward $\mathcal{R}^{(2)}$~(Immediate Reward).} At each step $t$, the LM policy produces a structured output
 in the reason-acting format~\citep{DBLP:conf/iclr/YaoZYDSN023}:
 \begin{equation}
    \small
     a_t=\texttt{<think>}...\texttt{</think><tool\_call>}~b_t~\texttt{</tool\_call>}, \nonumber
 \end{equation}
 where $b_t$ is the sub-action to perform policy-environment interaction by calling the specific scaffold, which follows the structured format:
 \begin{equation}
     \small
     a_t=\texttt{\{"name":...,"arguments":\{...\}\}}. \nonumber
 \end{equation}
 We set the reward to be $0.1$ if the response format is correct and the scaffold can be successfully parsed. Otherwise, the reward is $0$. We set $o^{(2)}=2$ and $gv^{(2)}=0$.

 \paragraph{Scaffold Calling Reward $\mathcal{R}^{(3)}$~(Immediate Reward).} For each scaffold calling sub-action $a_t$, if the parsed arguments are invalid, or the range of some specific argument is out-bounded, the scaffold will return error and the interaction will not be successfully executed. We set the reward to be $0.1$ if the scaffold are successfully executed, otherwise the reward will be $0$. We set $o^{(3)}=1$.

\paragraph{Scaffold Selection Reward $\mathcal{R}^{(4)}$~(Immediate Reward).} The reward function assigns different rewards for different scaffolds. Specifically, the reward is set to be $0.2$ if calling Shell, Editor and Submit scaffold. The reward is set to be $0.1$ if calling the Web Search scaffold. We set $o^{(4)}=1$.

The final reward will be the accumulation of different reward functions $\mathcal{R}^{(1)}$, $\mathcal{R}^{(2)}$, $\mathcal{R}^{(3)}$ and $\mathcal{R}^{(4)}$. Given the terminal state $s_T$ and action $a_T$, if $\mathcal{R}^{(1)}(s_T,a_T)\leq gv^{(1)}$, i.e., the patch is not successfully generated or the patch is empty, the stepwise critics $\mathcal{R}^{(2)}$, $\mathcal{R}^{(3)}$ and $\mathcal{R}^{(4)}$ will not be calculated, as their reward priority $o^{(2)},o^{(3)}$ and $o^{(4)}$ is lower than $o^{(1)}$. For each intermediate reasoning step with state $s_t$ and action $a_t$, if $\mathcal{R}^{(2)}\leq gv^{(2)}$, i.e., the sub-action $b_t$ is unable to be parsed, the scaffold calling reward $\mathcal{R}^{(3)}$ and the scaffold selection reward $\mathcal{R}^{(4)}$ will not be calculated.
\section{Experiment}
\subsection{Experiment Setup}

\paragraph{Baselines.} We compare the results of models including open-source models and proprietary models. For open-source models, we conduct our experiments on \texttt{Qwen2.5-3B-Instruct}~\citep{DBLP:journals/corr/abs-2412-15115}. we report results for the Supervised Fine-Tuned~(SFT) baseline, as well as models trained using RL with reward accumulation methods including our proposed G-RA method and the direct reward accumulation approach~(D-RA), which simply adds different rewards together. For proprietary models, we report the results of \texttt{Deepseek-V3}~\citep{DBLP:journals/corr/abs-2412-19437}, \texttt{Deepseek-R1}\citep{DBLP:journals/corr/abs-2501-12948}, \texttt{GPT-4o}~\citep{DBLP:journals/corr/abs-2410-21276} and \texttt{Gemini-2.5-Pro}~\citep{comanici2025gemini} as reference.

\paragraph{Training Details.} Our experiment involve multiple training stages. First of all, we perform SFT on Qwen for $2$ epoch. Then, we conduct RL training on the finetuned checkpoint for a maximum of $100$ steps. During the RL training stage, we optimize the policy model using the GRPO algorithm. More details are shown in Appendix~\ref{sec:app:training_details}.

\paragraph{Training Dataset} We perform post-training on SWE-bench-extra, a dataset that collect issue-pull request pairs from real Github issues. After filtering and data construction, we obtain $457$ docker images for trajectory generation and RL optimization. To construct the SFT dataset, we utilize \texttt{Deepseek-V3} for trajectory generation, and obtain $3$k trajectories in total. During the RL training process, we utilize task instances identical to those used to construct the SFT dataset. More details are shown in Appendix~\ref{sec:app:data_details}.

\paragraph{Evaluation Metrics} We evaluate the capabilities of all models in SWE-bench Verified~\citep{DBLP:conf/iclr/JimenezYWYPPN24} and a subset of kBench~\citep{DBLP:conf/nips/MathaiHMNIYR24}, named as kBench-50. We use the environment and scaffolds introduced in \S{\ref{sec:environment}} to perform experiments and evaluation. For all experiments, we report (1) Resolution Rate (RR), the proportion of resolved task instances, (2) Modification Rate (MR), the proportion of trajectories where the task repository is modified before calling the Submit scaffold, and (3) Completion Rate (CR), the proportion of trajectories where the Submit scaffold is called. More details are shown in Appendix~\ref{sec:app:evaluation}.

\subsection{Main Results}
We report the main results for both benchmarks in Table~\ref{tab:experiment-results}. Our observations are as follows: \textbf{(1) Both benchmarks are difficult for proprietary models and open-source models.} We observe a low resolution rate for proprietary models and open-source models. Deepseek-V3 outperforms other proprietary models and achieves a best performance in completion rate~($75.2\%$), modification rate~($62.6\%$) and resolution rate~($16.4\%$) in SWE-bench Verified. For other proprietary models, the resolution rates are less than $10\%$, showing that the task is difficult under current experiment settings. Especially, we observe that for Deepseek-R1 and Gemini-2.5-Pro, the completion rates are both lower than $20\%$, which indicate that the tasks are long-horizon and require multiple interactions between LM policy and the environment. For open-source models, the resolution rates are close to $0\%$ in most cases, showing that the tasks in both benchmarks are nearly impossible to be solved. In other words, the outcome rewards are sparse and the tasks instances are inefficient to provide valuable optimization feedback. \textbf{(2) Fine-grained Rewards can result in catastrophic policy degradation.} We observe a quick policy degradation when performing the D-RA method during the RL training process. Compared with the SFT baseline, the completion rate drops from $47.6\%$ to $19.4\%$ for the model trained with RL for $25$ steps when performing D-RA. The completion rate further drops to $1.4\%$ when training for $100$ steps, in which case the modification rate is $0.2\%$, showing that the LM policy makes almost no modifications to the task repositories during the multi-turn interaction before reaching the limitation of turn count. \textbf{(3) R-GA can effectively balance the sparse outcome reward and the dense stepwise critics. } We observe that the completion rate increases when performing RL training with G-RA method. Compared with the SFT baseline, the completion achieves an improvement of $46.2\%$, and the modification rate achieves an improvement of $2.8\%$ when training for $75$ steps. Compared with the D-RA method, G-RA mitigates the policy degradation observed when using direct reward accumulation method and stabilize RL optimization during the training process.

\subsection{Analysis and Discussion}
\begin{figure}[htbp]
    \centering
    \begin{subfigure}[b]{0.22\textwidth}
        \includegraphics[width=\textwidth]{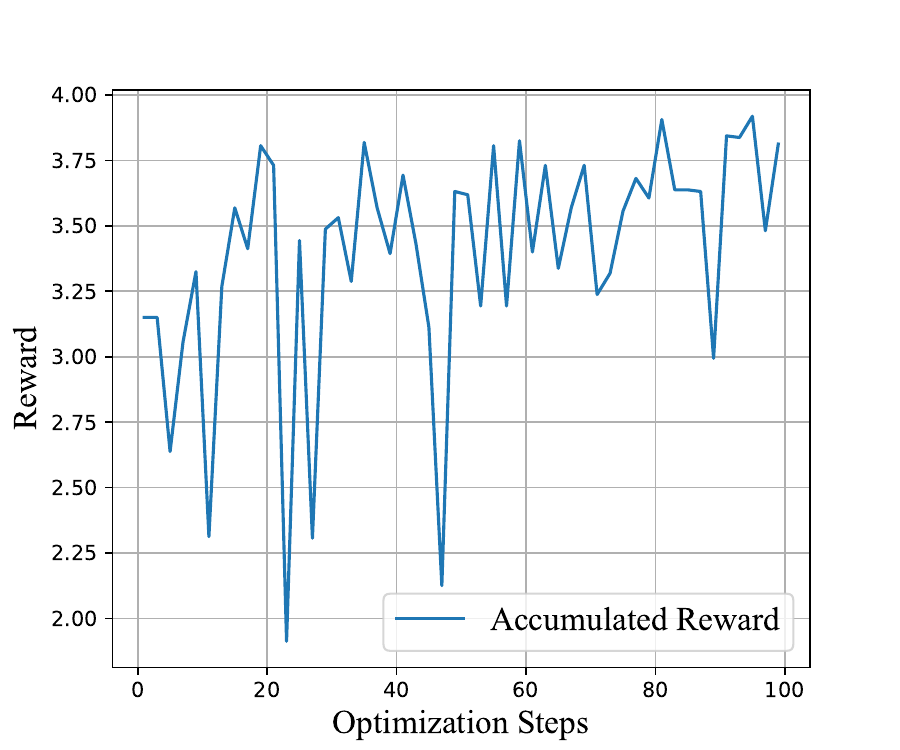}
        \caption{The accumulated reward.}
        \label{fig:mixed:accumulated}
    \end{subfigure}
    \hfill
    \begin{subfigure}[b]{0.22\textwidth}
        \includegraphics[width=\textwidth]{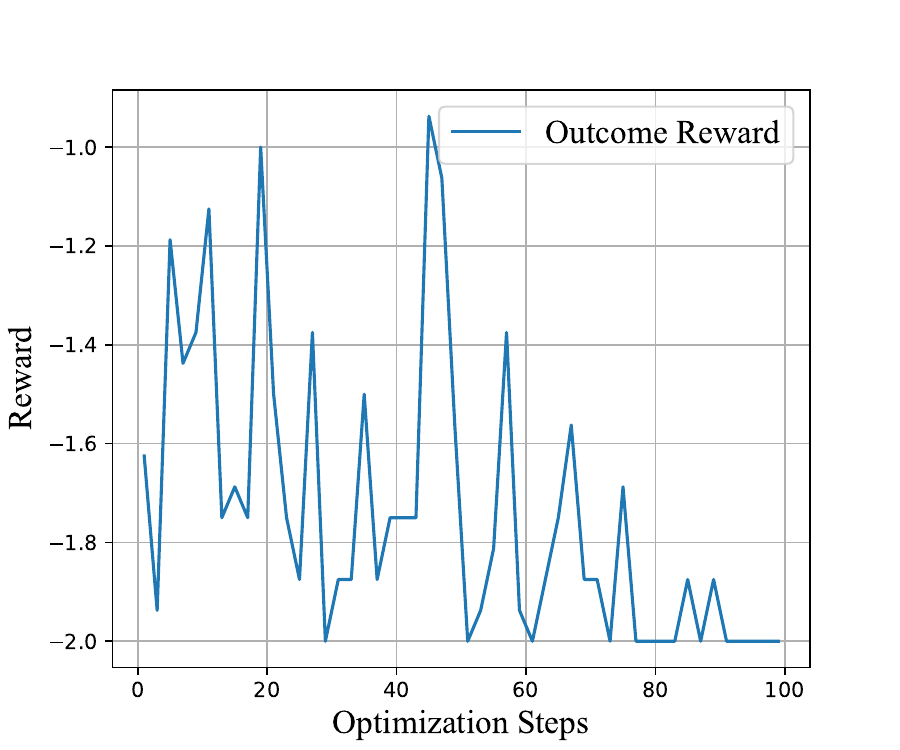}
        \caption{The outcome reward.}
        \label{fig:mixed:outcome}
    \end{subfigure}    
    \begin{subfigure}[b]{0.22\textwidth}
        \includegraphics[width=\textwidth]{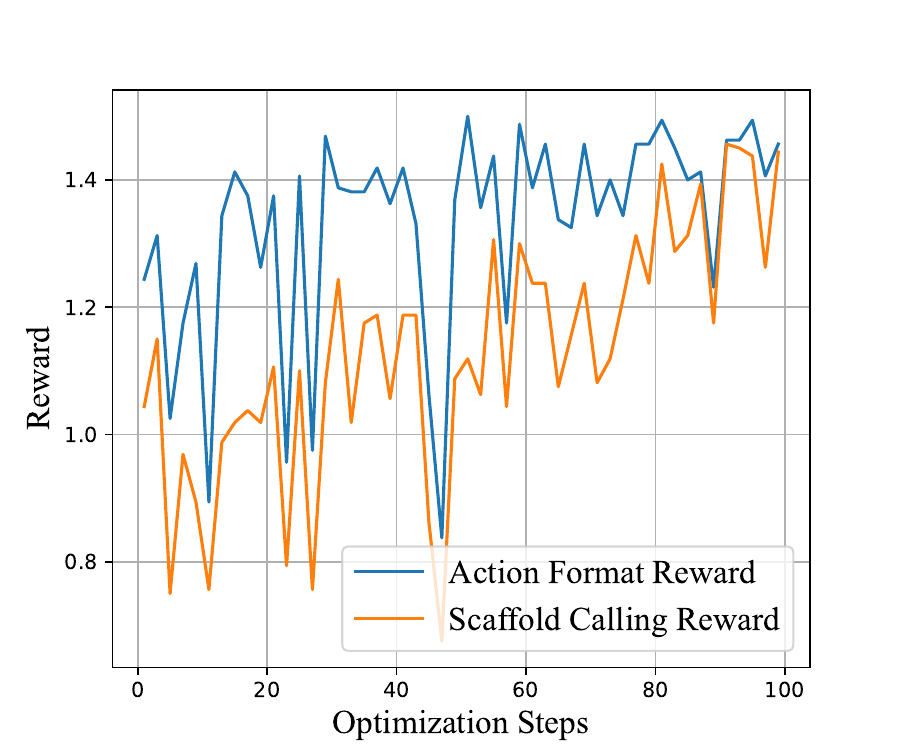}
        \caption{The action format reward and scaffold calling reward.}
        \label{fig:mixed:formattool}
    \end{subfigure}
    \hfill
    \begin{subfigure}[b]{0.22\textwidth}
        \includegraphics[width=\textwidth]{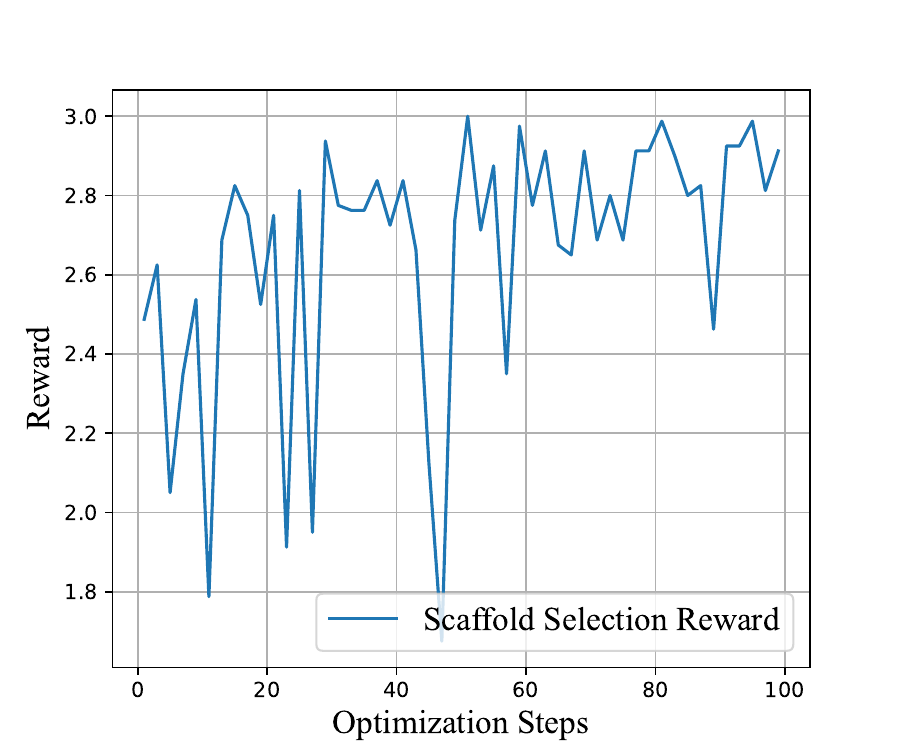}
        \caption{The scaffold selection reward.}
        \label{fig:mixed:action}
    \end{subfigure}
    
    \caption{The reward-Optimization Steps curve of RL training via D-RA. The accumulated reward keeps increasing, but the outcome reward decreases during RL process.}
    \label{fig:mixed_rewards}
\end{figure}

\paragraph{Policy Degradation.} We perform further analysis about the the catastrophic policy degradation phenomenon we observed when training open-source models with direct reward accumulation method during the RL process. We report the reward-optimization step curve in Figure~\ref{fig:mixed_rewards}, including the accumulated reward and its components, i.e., the outcome reward, the action format reward, the scaffold calling reward, and the scaffold selection reward. 

As shown in the results, although the accumulated reward keeps increasing during the RL training process~(Figure~\ref{fig:mixed:accumulated}), the increase is generally attributed to the verifiable immediate rewards~(Figure~\ref{fig:mixed:formattool} and Figure~\ref{fig:mixed:action}), while the outcome reward is negatively correlated with the accumulated reward (Figure~\ref{fig:mixed:outcome}). We attribute the phenomenon to reward-hacking, a problem that the LM policy learns to exploit flaws or unintended shortcuts in the reward function to maximize its reward in ways that do not align with the actual optimization goal, and is further due to the misalignment between different rule-based rewards: the long-term objective and the immediate rewards in long-horizon scenarios. We describe the misalignment in three folds. \textbf{(1) Granularity and Difficulty Misalignment.} The outcome reward is sparse and difficult to acquire. The outcome reward provides positive feedback when the problem is successfully handled, which is difficult for the open-source model. In other words, the outcome reward provides stable negative rewards during the RL training process, which provides ineffective signals towards the model optimization. Compared with the outcome reward, rewards derived from stepwise critics are dense and easy to acquire. \textbf{(2) Value Misalignment.} It is difficult to design the range of rewards, especially when the rewards are different in granularity and difficulty in multi-turn scenario. As shown in the results, the increase of immediate rewards results in the increase of the accumulated reward, as the decrease of outcome reward makes little influence to the tendency of the accumulated reward. \textbf{(3) Optimization Goal Misalignment.} The optimization goal of immediate rewards are not directly related to the long-term objective, as the verification-based rewards are heavily rely on human priors. This creates a divergence between the verifiable immediate rewards and the long-term objective, leading the LM policy to prioritize short-term gains over actual task completion. The results in Table~\ref{tab:experiment-results} also shows that during the RL training process, the long-term objective have gradually been neglected by the LM policy when training with D-RA, as the completion rate keeps decreasing.

\begin{figure}[htbp]
    \centering
    \begin{subfigure}[b]{0.22\textwidth}
        \includegraphics[width=\textwidth]{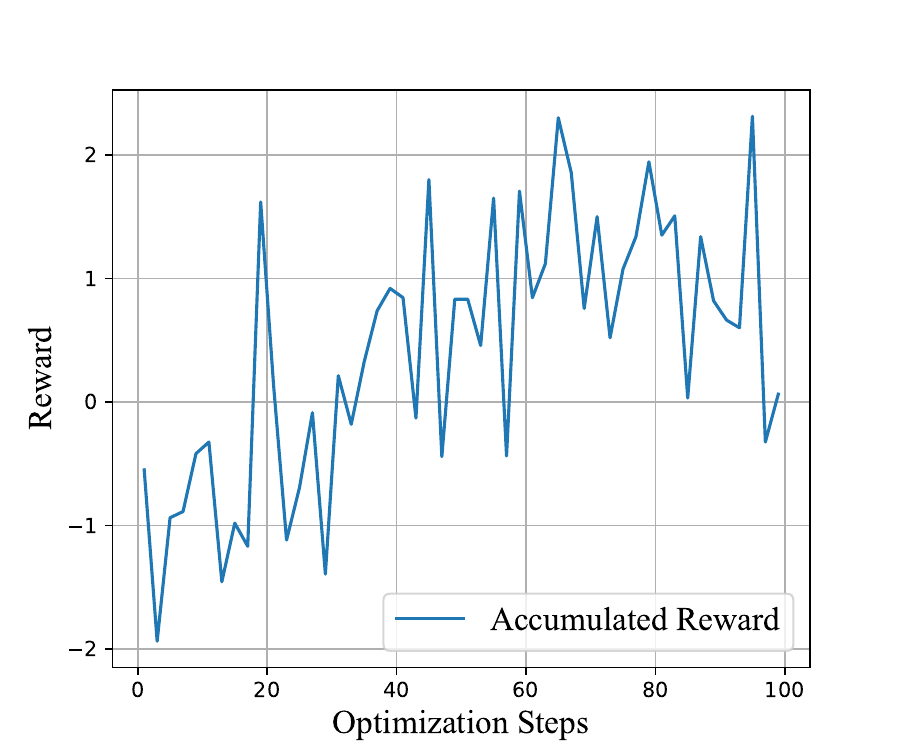}
        \caption{The accumulated reward.}
        \label{fig:ranked:accumulated}
    \end{subfigure}
    \hfill
    \begin{subfigure}[b]{0.22\textwidth}
        \includegraphics[width=\textwidth]{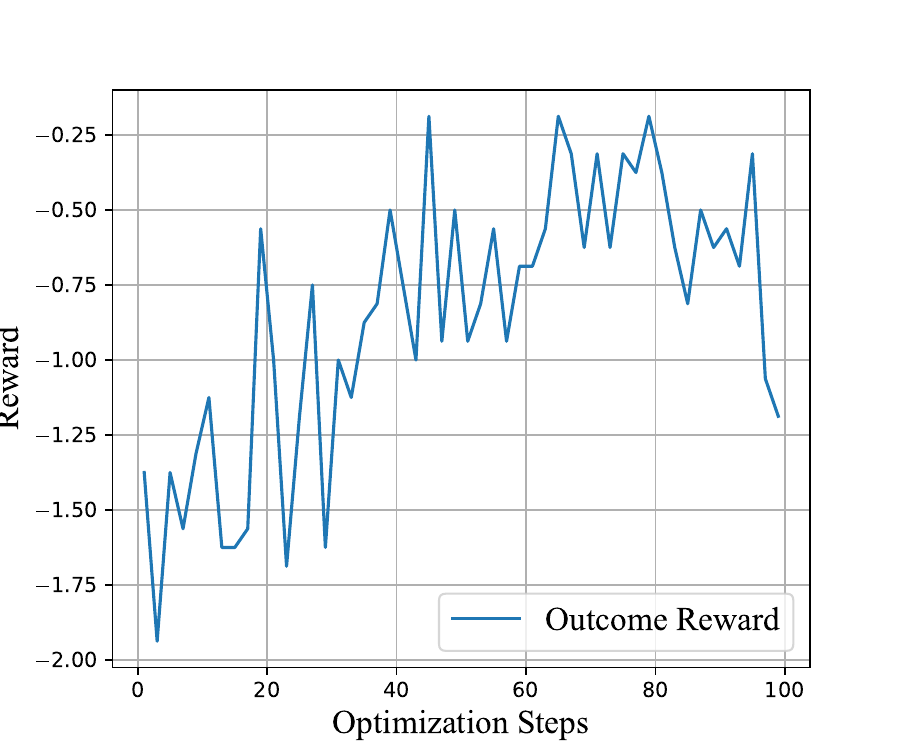}
        \caption{The outcome reward.}
        \label{fig:ranked:outcome}
    \end{subfigure}    
    \begin{subfigure}[b]{0.22\textwidth}
        \includegraphics[width=\textwidth]{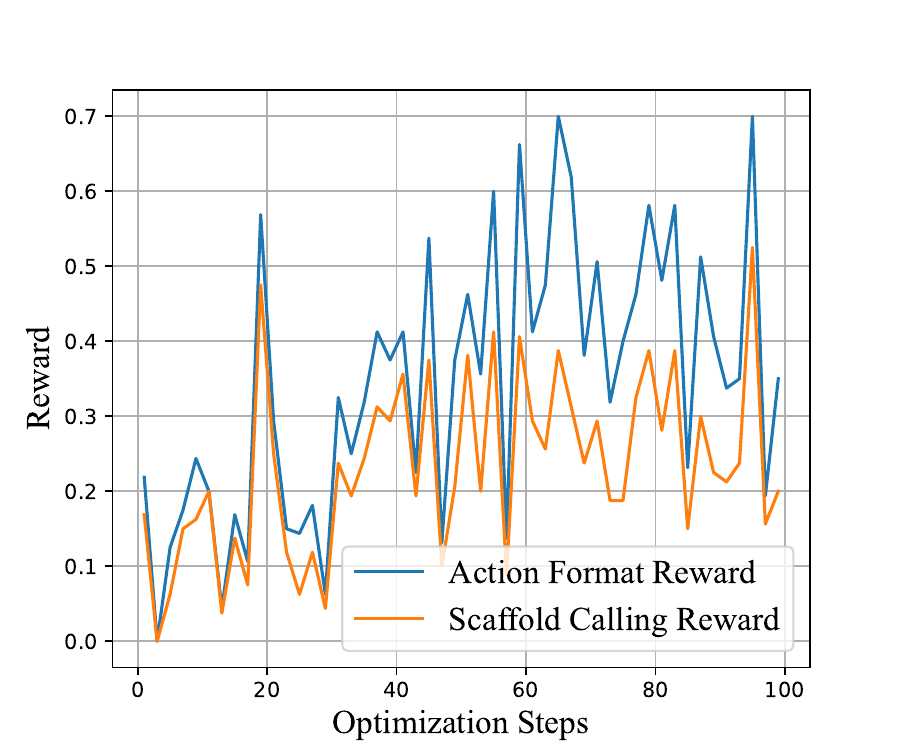}
        \caption{The action format reward and scaffold calling reward.}
        \label{fig:ranked:formattool}
    \end{subfigure}
    \hfill
    \begin{subfigure}[b]{0.22\textwidth}
        \includegraphics[width=\textwidth]{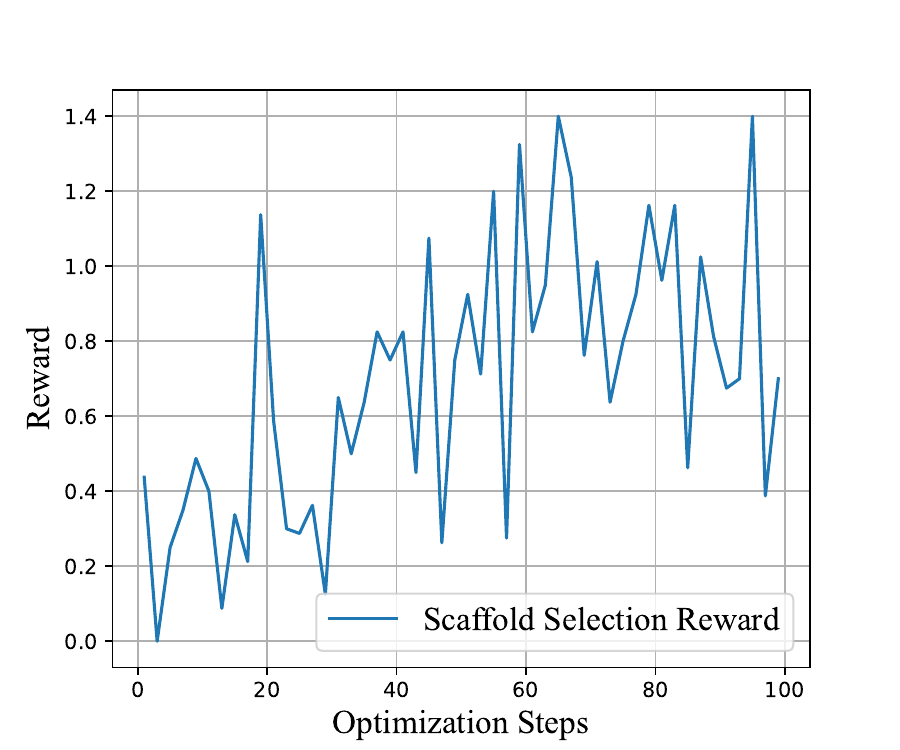}
        \caption{The scaffold selection reward.}
        \label{fig:ranked:action}
    \end{subfigure}
    
    \caption{The reward-Optimization Steps curve of RL training via G-RA. The rewards are continuously increasing.}
    \label{fig:ranked_rewards}
\end{figure}

In contrast, we provide the reward curve when the LM policy is trained with G-RA in Figure~\ref{fig:ranked_rewards}. Generally, G-RA prevents the LM policy from focusing on fitting stepwise critics at the cost of neglecting the long-term objective. As shown in the results, the outcome reward is increasing during the RL training process, which is correlated with the rewards derived from stepwise critics and the accumulated reward. The results highlight the importance of the careful reward design in long-horizon multi-turn scenario, as the fine-grained rewards can result in reward hacking problem and mislead the actual optimization goal. As a reward accumulation method, G-RA can effectively balance the different rewards, which is helpful for complex reasoning scenario.

\paragraph{Ablations of $gv^{(1)}$.} We further perform ablation study about the gate value $gv^{(1)}$, which controls whether the verifiable immediate reward will be neglected. We compare the results of model trained with G-RA varying on different $gv^{(1)}$ settings, including $gv^{(1)}=-2$~(not be neglected), $gv^{(1)}=-1$~(be neglected if LM fails to call the Submit scaffold), $gv^{(1)}=0$~( be neglected if LM generates an empty patch or fails to call the Submit scaffold), and $gv^{(1)}=10$~(be neglected if LM fails to solve the problem). 

\begin{figure}[htbp]
    \centering
    \begin{subfigure}[b]{0.22\textwidth}
        \includegraphics[width=\textwidth]{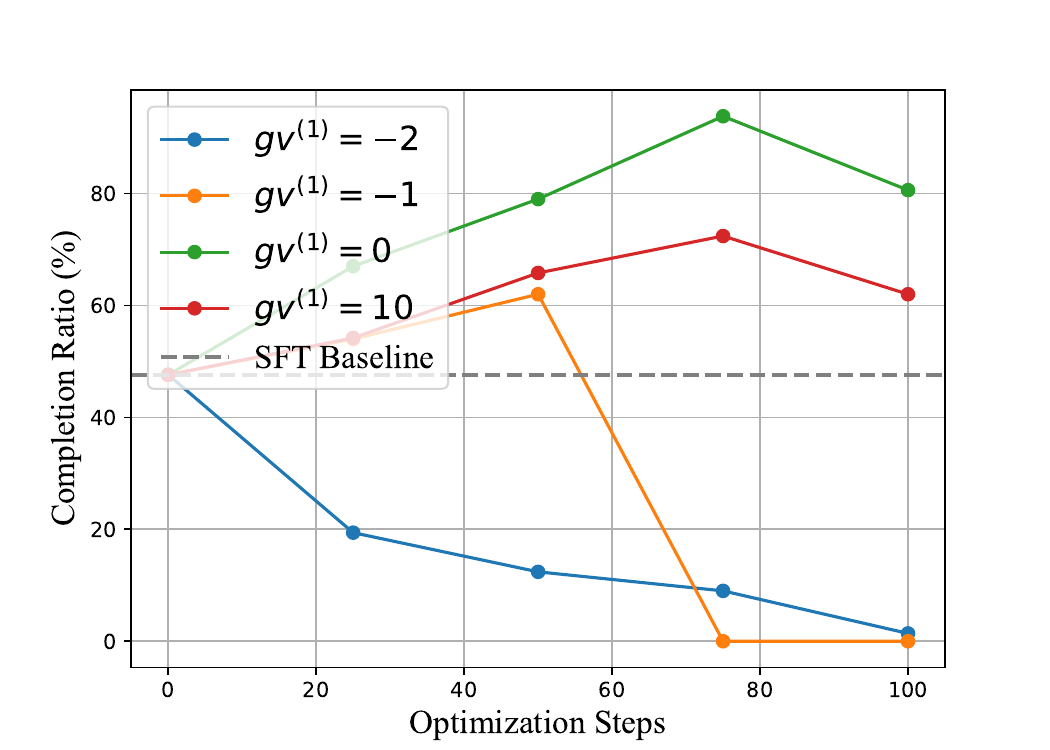}
        \caption{CR results.}
        \label{fig:ablation:gv_CR}
    \end{subfigure}
    \hfill
    \begin{subfigure}[b]{0.22\textwidth}
        \includegraphics[width=\textwidth]{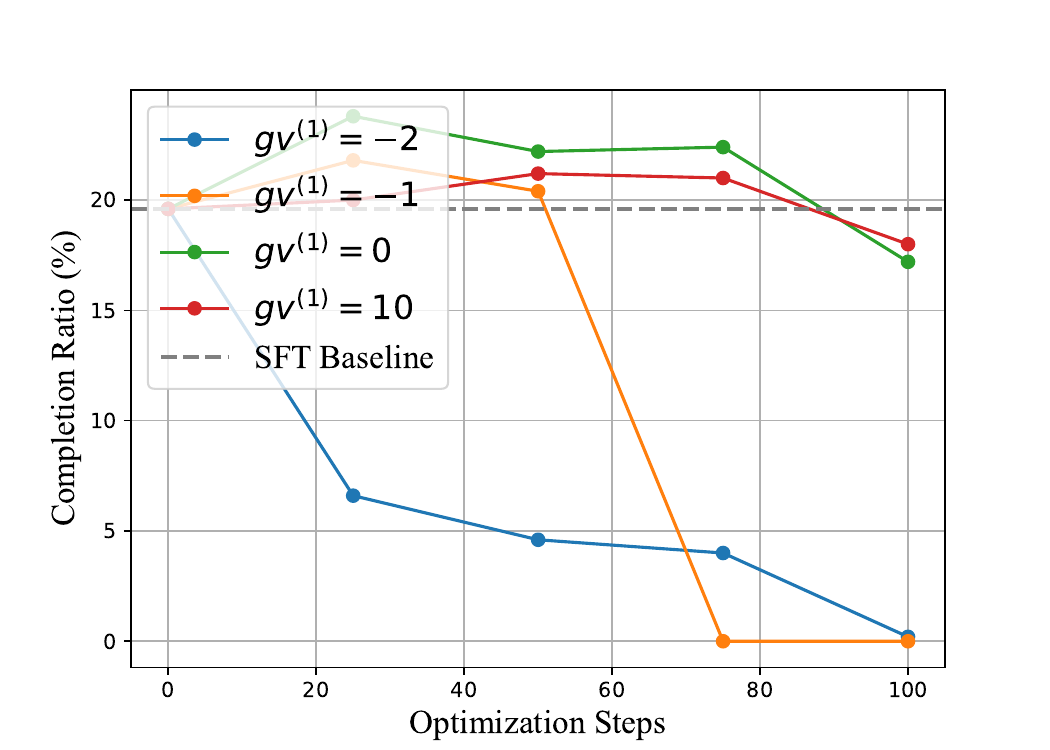}
        \caption{MR results.}
        \label{fig:ablation:gv_MR}
    \end{subfigure}    
    \caption{The CR and MR curve of models trained varying on different $gv^{(1)}$ settings.}
    \label{fig:ablation_gv}
\end{figure}

We report CR and MR in SWE-bench Verified in Figure~\ref{fig:ablation_gv}. The results show that lower gate values make verifiable immediate rewards easier to activate, but this also leads to more frequent policy degradation. For instance, when $gv^{(1)}=-2$, severe policy degradation occurs after $75$ optimization steps, reducing both CR and MR to zero. In contrast, setting $gv^{(1)}=10$ minimizes the influence of verifiable immediate rewards, causing optimization to rely primarily on the long-horizon objective. Compared to $gv^{(1)}=-1$, this setting results in slower improvements in CR and MR, suggesting that immediate rewards can be useful for long-horizon RL.

\paragraph{Echo Trap.} \cite{DBLP:journals/corr/abs-2504-20073} introduces the ``echo trap" phenomenon during the multi-turn RL optimization: the LM policy overfits to locally rewarded reasoning patterns, which is a consequence of the reward-hacking problem and the unreasonable accumulation of the verifiable immediate rewards and the long-term objectives. In our experiments, we observe the similar phenomenon for the LM policy trained with V-RA. Specifically, the model repeatedly uses the Shell scaffold and the ``grep" command until the turn number reaches the upper bound. These operations will not modify the repository or submit the modification to end the task, but will accumulate the verifiable immediate rewards. We provide examples with different failure modes in Appendix~\ref{sec:app:cases}.
\section{Conclusion}
We propose Gated Reward Accumulation (G-RA), a novel reward accumulation method that stabilizes RL optimization in long-horizon tasks by conditioning intermediate rewards on long-term objective achievement. Integrated with our SWE-oriented RL Framework, G-RA improves completion rates and modification rates while preventing reward misalignment. This work provides a practical solution for stabilizing RL training in complex reasoning domains requiring multi-turn interaction and verifiable rewards, especially when the long-term objective is sparse and struggling.

\bibliography{aaai}

\begin{thebibliography}{34}
\providecommand{\natexlab}[1]{#1}

\bibitem[{Brockman et~al.(2016)Brockman, Cheung, Pettersson, Schneider, Schulman, Tang, and Zaremba}]{brockman2016openai}
Brockman, G.; Cheung, V.; Pettersson, L.; Schneider, J.; Schulman, J.; Tang, J.; and Zaremba, W. 2016.
\newblock Openai gym.
\newblock \emph{arXiv preprint arXiv:1606.01540}.

\bibitem[{Comanici et~al.(2025)Comanici, Bieber, Schaekermann, Pasupat, Sachdeva, Dhillon, Blistein, Ram, Zhang, Rosen et~al.}]{comanici2025gemini}
Comanici, G.; Bieber, E.; Schaekermann, M.; Pasupat, I.; Sachdeva, N.; Dhillon, I.; Blistein, M.; Ram, O.; Zhang, D.; Rosen, E.; et~al. 2025.
\newblock Gemini 2.5: Pushing the frontier with advanced reasoning, multimodality, long context, and next generation agentic capabilities.
\newblock \emph{arXiv preprint arXiv:2507.06261}.

\bibitem[{DeepSeek{-}AI et~al.(2025)DeepSeek{-}AI, Guo, Yang, Zhang, Song, Zhang, Xu, Zhu, Ma, Wang, Bi, Zhang, Yu, Wu, Wu, Gou, Shao, Li, Gao, Liu, Xue, Wang, Wu, Feng, Lu, Zhao, Deng, Zhang, Ruan, Dai, Chen, Ji, Li, Lin, Dai, Luo, Hao, Chen, Li, Zhang, Bao, Xu, Wang, Ding, Xin, Gao, Qu, Li, Guo, Li, Wang, Chen, Yuan, Qiu, Li, Cai, Ni, Liang, Chen, Dong, Hu, Gao, Guan, Huang, Yu, Wang, Zhang, Zhao, Wang, Zhang, Xu, Xia, Zhang, Zhang, Tang, Li, Wang, Li, Tian, Huang, Zhang, Wang, Chen, Du, Ge, Zhang, Pan, Wang, Chen, Jin, Chen, Lu, Zhou, Chen, Ye, Wang, Yu, Zhou, Pan, and Li}]{DBLP:journals/corr/abs-2501-12948}
DeepSeek{-}AI; Guo, D.; Yang, D.; Zhang, H.; Song, J.; Zhang, R.; Xu, R.; Zhu, Q.; Ma, S.; Wang, P.; Bi, X.; Zhang, X.; Yu, X.; Wu, Y.; Wu, Z.~F.; Gou, Z.; Shao, Z.; Li, Z.; Gao, Z.; Liu, A.; Xue, B.; Wang, B.; Wu, B.; Feng, B.; Lu, C.; Zhao, C.; Deng, C.; Zhang, C.; Ruan, C.; Dai, D.; Chen, D.; Ji, D.; Li, E.; Lin, F.; Dai, F.; Luo, F.; Hao, G.; Chen, G.; Li, G.; Zhang, H.; Bao, H.; Xu, H.; Wang, H.; Ding, H.; Xin, H.; Gao, H.; Qu, H.; Li, H.; Guo, J.; Li, J.; Wang, J.; Chen, J.; Yuan, J.; Qiu, J.; Li, J.; Cai, J.~L.; Ni, J.; Liang, J.; Chen, J.; Dong, K.; Hu, K.; Gao, K.; Guan, K.; Huang, K.; Yu, K.; Wang, L.; Zhang, L.; Zhao, L.; Wang, L.; Zhang, L.; Xu, L.; Xia, L.; Zhang, M.; Zhang, M.; Tang, M.; Li, M.; Wang, M.; Li, M.; Tian, N.; Huang, P.; Zhang, P.; Wang, Q.; Chen, Q.; Du, Q.; Ge, R.; Zhang, R.; Pan, R.; Wang, R.; Chen, R.~J.; Jin, R.~L.; Chen, R.; Lu, S.; Zhou, S.; Chen, S.; Ye, S.; Wang, S.; Yu, S.; Zhou, S.; Pan, S.; and Li, S.~S. 2025.
\newblock DeepSeek-R1: Incentivizing Reasoning Capability in LLMs via Reinforcement Learning.
\newblock \emph{CoRR}, abs/2501.12948.

\bibitem[{DeepSeek{-}AI et~al.(2024)DeepSeek{-}AI, Liu, Feng, Xue, Wang, Wu, Lu, Zhao, Deng, Zhang, Ruan, Dai, Guo, Yang, Chen, Ji, Li, Lin, Dai, Luo, Hao, Chen, Li, Zhang, Bao, Xu, Wang, Zhang, Ding, Xin, Gao, Li, Qu, Cai, Liang, Guo, Ni, Li, Wang, Chen, Chen, Yuan, Qiu, Li, Song, Dong, Hu, Gao, Guan, Huang, Yu, Wang, Zhang, Xu, Xia, Zhao, Wang, Zhang, Li, Wang, Zhang, Zhang, Tang, Li, Tian, Huang, Wang, Zhang, Wang, Zhu, Chen, Du, Chen, Jin, Ge, Zhang, Pan, Wang, Xu, Zhang, Chen, Li, Lu, Zhou, Chen, Wu, Ye, Ye, Ma, Wang, Zhou, Yu, Zhou, Pan, Wang, Yun, Pei, Sun, Xiao, and Zeng}]{DBLP:journals/corr/abs-2412-19437}
DeepSeek{-}AI; Liu, A.; Feng, B.; Xue, B.; Wang, B.; Wu, B.; Lu, C.; Zhao, C.; Deng, C.; Zhang, C.; Ruan, C.; Dai, D.; Guo, D.; Yang, D.; Chen, D.; Ji, D.; Li, E.; Lin, F.; Dai, F.; Luo, F.; Hao, G.; Chen, G.; Li, G.; Zhang, H.; Bao, H.; Xu, H.; Wang, H.; Zhang, H.; Ding, H.; Xin, H.; Gao, H.; Li, H.; Qu, H.; Cai, J.~L.; Liang, J.; Guo, J.; Ni, J.; Li, J.; Wang, J.; Chen, J.; Chen, J.; Yuan, J.; Qiu, J.; Li, J.; Song, J.; Dong, K.; Hu, K.; Gao, K.; Guan, K.; Huang, K.; Yu, K.; Wang, L.; Zhang, L.; Xu, L.; Xia, L.; Zhao, L.; Wang, L.; Zhang, L.; Li, M.; Wang, M.; Zhang, M.; Zhang, M.; Tang, M.; Li, M.; Tian, N.; Huang, P.; Wang, P.; Zhang, P.; Wang, Q.; Zhu, Q.; Chen, Q.; Du, Q.; Chen, R.~J.; Jin, R.~L.; Ge, R.; Zhang, R.; Pan, R.; Wang, R.; Xu, R.; Zhang, R.; Chen, R.; Li, S.~S.; Lu, S.; Zhou, S.; Chen, S.; Wu, S.; Ye, S.; Ye, S.; Ma, S.; Wang, S.; Zhou, S.; Yu, S.; Zhou, S.; Pan, S.; Wang, T.; Yun, T.; Pei, T.; Sun, T.; Xiao, W.~L.; and Zeng, W. 2024.
\newblock DeepSeek-V3 Technical Report.
\newblock \emph{CoRR}, abs/2412.19437.

\bibitem[{Dietterich(2000)}]{DBLP:journals/jair/Dietterich00}
Dietterich, T.~G. 2000.
\newblock Hierarchical Reinforcement Learning with the {MAXQ} Value Function Decomposition.
\newblock \emph{J. Artif. Intell. Res.}, 13: 227--303.

\bibitem[{Du et~al.(2024)Du, Li, Torralba, Tenenbaum, and Mordatch}]{DBLP:conf/icml/Du00TM24}
Du, Y.; Li, S.; Torralba, A.; Tenenbaum, J.~B.; and Mordatch, I. 2024.
\newblock Improving Factuality and Reasoning in Language Models through Multiagent Debate.
\newblock In \emph{{ICML}}. OpenReview.net.

\bibitem[{Forbes and Roberts(2024)}]{DBLP:conf/aaai/Forbes024}
Forbes, G.~C.; and Roberts, D.~L. 2024.
\newblock Potential-Based Reward Shaping for Intrinsic Motivation (Student Abstract).
\newblock In \emph{{AAAI}}, 23488--23489. {AAAI} Press.

\bibitem[{Gupta et~al.(2023)Gupta, Chandak, Jordan, Thomas, and C~da Silva}]{gupta2023behavior}
Gupta, D.; Chandak, Y.; Jordan, S.; Thomas, P.~S.; and C~da Silva, B. 2023.
\newblock Behavior alignment via reward function optimization.
\newblock \emph{Advances in Neural Information Processing Systems}, 36: 52759--52791.

\bibitem[{Hengst(2010)}]{DBLP:reference/ml/Hengst10}
Hengst, B. 2010.
\newblock Hierarchical Reinforcement Learning.
\newblock In \emph{Encyclopedia of Machine Learning}, 495--502. Springer.

\bibitem[{Hu et~al.(2020)Hu, Wang, Jia, Wang, Chen, Hao, Wu, and Fan}]{DBLP:conf/nips/HuWJWCH0F20}
Hu, Y.; Wang, W.; Jia, H.; Wang, Y.; Chen, Y.; Hao, J.; Wu, F.; and Fan, C. 2020.
\newblock Learning to Utilize Shaping Rewards: {A} New Approach of Reward Shaping.
\newblock In \emph{NeurIPS}.

\bibitem[{Hurst et~al.(2024)Hurst, Lerer, Goucher, Perelman, Ramesh, Clark, Ostrow, Welihinda, Hayes, Radford, Madry, Baker{-}Whitcomb, Beutel, Borzunov, Carney, Chow, Kirillov, Nichol, Paino, Renzin, Passos, Kirillov, Christakis, Conneau, Kamali, Jabri, Moyer, Tam, Crookes, Tootoonchian, Kumar, Vallone, Karpathy, Braunstein, Cann, Codispoti, Galu, Kondrich, Tulloch, Mishchenko, Baek, Jiang, Pelisse, Woodford, Gosalia, Dhar, Pantuliano, Nayak, Oliver, Zoph, Ghorbani, Leimberger, Rossen, Sokolowsky, Wang, Zweig, Hoover, Samic, McGrew, Spero, Giertler, Cheng, Lightcap, Walkin, Quinn, Guarraci, Hsu, Kellogg, Eastman, Lugaresi, Wainwright, Bassin, Hudson, Chu, Nelson, Li, Shern, Conger, Barette, Voss, Ding, Lu, Zhang, Beaumont, Hallacy, Koch, Gibson, Kim, Choi, McLeavey, Hesse, Fischer, Winter, Czarnecki, Jarvis, Wei, Koumouzelis, and Sherburn}]{DBLP:journals/corr/abs-2410-21276}
Hurst, A.; Lerer, A.; Goucher, A.~P.; Perelman, A.; Ramesh, A.; Clark, A.; Ostrow, A.; Welihinda, A.; Hayes, A.; Radford, A.; Madry, A.; Baker{-}Whitcomb, A.; Beutel, A.; Borzunov, A.; Carney, A.; Chow, A.; Kirillov, A.; Nichol, A.; Paino, A.; Renzin, A.; Passos, A.~T.; Kirillov, A.; Christakis, A.; Conneau, A.; Kamali, A.; Jabri, A.; Moyer, A.; Tam, A.; Crookes, A.; Tootoonchian, A.; Kumar, A.; Vallone, A.; Karpathy, A.; Braunstein, A.; Cann, A.; Codispoti, A.; Galu, A.; Kondrich, A.; Tulloch, A.; Mishchenko, A.; Baek, A.; Jiang, A.; Pelisse, A.; Woodford, A.; Gosalia, A.; Dhar, A.; Pantuliano, A.; Nayak, A.; Oliver, A.; Zoph, B.; Ghorbani, B.; Leimberger, B.; Rossen, B.; Sokolowsky, B.; Wang, B.; Zweig, B.; Hoover, B.; Samic, B.; McGrew, B.; Spero, B.; Giertler, B.; Cheng, B.; Lightcap, B.; Walkin, B.; Quinn, B.; Guarraci, B.; Hsu, B.; Kellogg, B.; Eastman, B.; Lugaresi, C.; Wainwright, C.~L.; Bassin, C.; Hudson, C.; Chu, C.; Nelson, C.; Li, C.; Shern, C.~J.; Conger, C.; Barette, C.; Voss, C.; Ding, C.; Lu,
  C.; Zhang, C.; Beaumont, C.; Hallacy, C.; Koch, C.; Gibson, C.; Kim, C.; Choi, C.; McLeavey, C.; Hesse, C.; Fischer, C.; Winter, C.; Czarnecki, C.; Jarvis, C.; Wei, C.; Koumouzelis, C.; and Sherburn, D. 2024.
\newblock GPT-4o System Card.
\newblock \emph{CoRR}, abs/2410.21276.

\bibitem[{Ibrahim et~al.(2024)Ibrahim, Mostafa, Jnadi, Salloum, and Osinenko}]{DBLP:journals/access/IbrahimMJSO24}
Ibrahim, S.; Mostafa, M.; Jnadi, A.; Salloum, H.; and Osinenko, P. 2024.
\newblock Comprehensive Overview of Reward Engineering and Shaping in Advancing Reinforcement Learning Applications.
\newblock \emph{{IEEE} Access}, 12: 175473--175500.

\bibitem[{Jimenez et~al.(2024)Jimenez, Yang, Wettig, Yao, Pei, Press, and Narasimhan}]{DBLP:conf/iclr/JimenezYWYPPN24}
Jimenez, C.~E.; Yang, J.; Wettig, A.; Yao, S.; Pei, K.; Press, O.; and Narasimhan, K.~R. 2024.
\newblock SWE-bench: Can Language Models Resolve Real-world Github Issues?
\newblock In \emph{{ICLR}}. OpenReview.net.

\bibitem[{Li et~al.(2023)Li, Hammoud, Itani, Khizbullin, and Ghanem}]{DBLP:conf/nips/LiHIKG23}
Li, G.; Hammoud, H.; Itani, H.; Khizbullin, D.; and Ghanem, B. 2023.
\newblock {CAMEL:} Communicative Agents for "Mind" Exploration of Large Language Model Society.
\newblock In \emph{NeurIPS}.

\bibitem[{Mathai et~al.(2024)Mathai, Huang, Maniatis, Nogikh, Ivancic, Yang, and Ray}]{DBLP:conf/nips/MathaiHMNIYR24}
Mathai, A.; Huang, C.; Maniatis, P.; Nogikh, A.; Ivancic, F.; Yang, J.; and Ray, B. 2024.
\newblock kGym: {A} Platform and Dataset to Benchmark Large Language Models on Linux Kernel Crash Resolution.
\newblock In \emph{NeurIPS}.

\bibitem[{Nayyar and Srivastava(2025)}]{DBLP:conf/aaai/Nayyar025}
Nayyar, R.~K.; and Srivastava, S. 2025.
\newblock Autonomous Option Invention for Continual Hierarchical Reinforcement Learning and Planning.
\newblock In \emph{{AAAI}}, 19642--19650. {AAAI} Press.

\bibitem[{Ng, Harada, and Russell(1999)}]{DBLP:conf/icml/NgHR99}
Ng, A.~Y.; Harada, D.; and Russell, S. 1999.
\newblock Policy Invariance Under Reward Transformations: Theory and Application to Reward Shaping.
\newblock In \emph{{ICML}}, 278--287. Morgan Kaufmann.

\bibitem[{Ranchod, Rosman, and Konidaris(2015)}]{DBLP:conf/iros/RanchodRK15}
Ranchod, P.; Rosman, B.; and Konidaris, G.~D. 2015.
\newblock Nonparametric Bayesian reward segmentation for skill discovery using inverse reinforcement learning.
\newblock In \emph{2015 {IEEE/RSJ} International Conference on Intelligent Robots and Systems, {IROS} 2015, Hamburg, Germany, September 28 - October 2, 2015}, 471--477. {IEEE}.

\bibitem[{Shao et~al.(2024)Shao, Wang, Zhu, Xu, Song, Zhang, Li, Wu, and Guo}]{DBLP:journals/corr/abs-2402-03300}
Shao, Z.; Wang, P.; Zhu, Q.; Xu, R.; Song, J.; Zhang, M.; Li, Y.~K.; Wu, Y.; and Guo, D. 2024.
\newblock DeepSeekMath: Pushing the Limits of Mathematical Reasoning in Open Language Models.
\newblock \emph{CoRR}, abs/2402.03300.

\bibitem[{Sharma et~al.(2019)Sharma, Sharma, Rhinehart, and Kitani}]{DBLP:conf/iclr/SharmaSRK19}
Sharma, M.; Sharma, A.; Rhinehart, N.; and Kitani, K.~M. 2019.
\newblock Directed-Info {GAIL:} Learning Hierarchical Policies from Unsegmented Demonstrations using Directed Information.
\newblock In \emph{7th International Conference on Learning Representations, {ICLR} 2019, New Orleans, LA, USA, May 6-9, 2019}. OpenReview.net.

\bibitem[{Sheng et~al.(2025)Sheng, Zhang, Ye, Wu, Zhang, Zhang, Peng, Lin, and Wu}]{DBLP:conf/eurosys/ShengZYWZZPL025}
Sheng, G.; Zhang, C.; Ye, Z.; Wu, X.; Zhang, W.; Zhang, R.; Peng, Y.; Lin, H.; and Wu, C. 2025.
\newblock HybridFlow: {A} Flexible and Efficient {RLHF} Framework.
\newblock In \emph{EuroSys}, 1279--1297. {ACM}.

\bibitem[{Wang et~al.(2024)Wang, Wang, Su, Tong, and Song}]{DBLP:conf/acl/WangWSTS24}
Wang, Q.; Wang, Z.; Su, Y.; Tong, H.; and Song, Y. 2024.
\newblock Rethinking the Bounds of {LLM} Reasoning: Are Multi-Agent Discussions the Key?
\newblock In \emph{{ACL} {(1)}}, 6106--6131. Association for Computational Linguistics.

\bibitem[{Wang et~al.(2025{\natexlab{a}})Wang, Li, Song, Xu, Tang, Zhuge, Pan, Song, Li, Singh, Tran, Li, Ma, Zheng, Qian, Shao, Muennighoff, Zhang, Hui, Lin, and et~al.}]{DBLP:conf/iclr/0001LSXTZPSLSTL25}
Wang, X.; Li, B.; Song, Y.; Xu, F.~F.; Tang, X.; Zhuge, M.; Pan, J.; Song, Y.; Li, B.; Singh, J.; Tran, H.~H.; Li, F.; Ma, R.; Zheng, M.; Qian, B.; Shao, Y.; Muennighoff, N.; Zhang, Y.; Hui, B.; Lin, J.; and et~al. 2025{\natexlab{a}}.
\newblock OpenHands: An Open Platform for {AI} Software Developers as Generalist Agents.
\newblock In \emph{The Thirteenth International Conference on Learning Representations, {ICLR} 2025, Singapore, April 24-28, 2025}. OpenReview.net.

\bibitem[{Wang et~al.(2023)Wang, Yang, Dong, Sun, Liu, and U}]{DBLP:conf/nips/WangYDSLU23}
Wang, Y.; Yang, M.; Dong, R.; Sun, B.; Liu, F.; and U, L.~H. 2023.
\newblock Efficient Potential-based Exploration in Reinforcement Learning using Inverse Dynamic Bisimulation Metric.
\newblock In \emph{NeurIPS}.

\bibitem[{Wang et~al.(2025{\natexlab{b}})Wang, Yang, Zeng, Ren, Liu, Peng, Cheng, He, Wang, Gao, Chen, Wang, Du, and Shen}]{DBLP:journals/corr/abs-2504-20571}
Wang, Y.; Yang, Q.; Zeng, Z.; Ren, L.; Liu, L.; Peng, B.; Cheng, H.; He, X.; Wang, K.; Gao, J.; Chen, W.; Wang, S.; Du, S.~S.; and Shen, Y. 2025{\natexlab{b}}.
\newblock Reinforcement Learning for Reasoning in Large Language Models with One Training Example.
\newblock \emph{CoRR}, abs/2504.20571.

\bibitem[{Wang et~al.(2025{\natexlab{c}})Wang, Wang, Wang, Zhang, Li, Yang, Jin, Yu, Nguyen, Liu, Gottlieb, Lu, Cho, Wu, Fei{-}Fei, Wang, Choi, and Li}]{DBLP:journals/corr/abs-2504-20073}
Wang, Z.; Wang, K.; Wang, Q.; Zhang, P.; Li, L.; Yang, Z.; Jin, X.; Yu, K.; Nguyen, M.~N.; Liu, L.; Gottlieb, E.; Lu, Y.; Cho, K.; Wu, J.; Fei{-}Fei, L.; Wang, L.; Choi, Y.; and Li, M. 2025{\natexlab{c}}.
\newblock {RAGEN:} Understanding Self-Evolution in {LLM} Agents via Multi-Turn Reinforcement Learning.
\newblock \emph{CoRR}, abs/2504.20073.

\bibitem[{Xie et~al.(2024)Xie, Zhang, Chen, Li, Zhao, Cao, Hua, Cheng, Shin, Lei, Liu, Xu, Zhou, Savarese, Xiong, Zhong, and Yu}]{DBLP:conf/nips/XieZCLZCHCSLLXZ24}
Xie, T.; Zhang, D.; Chen, J.; Li, X.; Zhao, S.; Cao, R.; Hua, T.~J.; Cheng, Z.; Shin, D.; Lei, F.; Liu, Y.; Xu, Y.; Zhou, S.; Savarese, S.; Xiong, C.; Zhong, V.; and Yu, T. 2024.
\newblock OSWorld: Benchmarking Multimodal Agents for Open-Ended Tasks in Real Computer Environments.
\newblock In \emph{NeurIPS}.

\bibitem[{Yang et~al.(2024{\natexlab{a}})Yang, Yang, Zhang, Hui, Zheng, Yu, Li, Liu, Huang, Wei, Lin, Yang, Tu, Zhang, Yang, Yang, Zhou, Lin, Dang, Lu, Bao, Yang, Yu, Li, Xue, Zhang, Zhu, Men, Lin, Li, Xia, Ren, Ren, Fan, Su, Zhang, Wan, Liu, Cui, Zhang, and Qiu}]{DBLP:journals/corr/abs-2412-15115}
Yang, A.; Yang, B.; Zhang, B.; Hui, B.; Zheng, B.; Yu, B.; Li, C.; Liu, D.; Huang, F.; Wei, H.; Lin, H.; Yang, J.; Tu, J.; Zhang, J.; Yang, J.; Yang, J.; Zhou, J.; Lin, J.; Dang, K.; Lu, K.; Bao, K.; Yang, K.; Yu, L.; Li, M.; Xue, M.; Zhang, P.; Zhu, Q.; Men, R.; Lin, R.; Li, T.; Xia, T.; Ren, X.; Ren, X.; Fan, Y.; Su, Y.; Zhang, Y.; Wan, Y.; Liu, Y.; Cui, Z.; Zhang, Z.; and Qiu, Z. 2024{\natexlab{a}}.
\newblock Qwen2.5 Technical Report.
\newblock \emph{CoRR}, abs/2412.15115.

\bibitem[{Yang et~al.(2024{\natexlab{b}})Yang, Jimenez, Wettig, Lieret, Yao, Narasimhan, and Press}]{DBLP:conf/nips/YangJWLYNP24}
Yang, J.; Jimenez, C.~E.; Wettig, A.; Lieret, K.; Yao, S.; Narasimhan, K.; and Press, O. 2024{\natexlab{b}}.
\newblock SWE-agent: Agent-Computer Interfaces Enable Automated Software Engineering.
\newblock In Globersons, A.; Mackey, L.; Belgrave, D.; Fan, A.; Paquet, U.; Tomczak, J.~M.; and Zhang, C., eds., \emph{Advances in Neural Information Processing Systems 38: Annual Conference on Neural Information Processing Systems 2024, NeurIPS 2024, Vancouver, BC, Canada, December 10 - 15, 2024}.

\bibitem[{Yao et~al.(2025)Yao, Shinn, Razavi, and Narasimhan}]{DBLP:conf/iclr/YaoSRN25}
Yao, S.; Shinn, N.; Razavi, P.; and Narasimhan, K.~R. 2025.
\newblock \{{\(\tau\)}\}-bench: {A} Benchmark for {\textbackslash}underline\{T\}ool-{\textbackslash}underline\{A\}gent-{\textbackslash}underline\{U\}ser Interaction in Real-World Domains.
\newblock In \emph{{ICLR}}. OpenReview.net.

\bibitem[{Yao et~al.(2023)Yao, Zhao, Yu, Du, Shafran, Narasimhan, and Cao}]{DBLP:conf/iclr/YaoZYDSN023}
Yao, S.; Zhao, J.; Yu, D.; Du, N.; Shafran, I.; Narasimhan, K.~R.; and Cao, Y. 2023.
\newblock ReAct: Synergizing Reasoning and Acting in Language Models.
\newblock In \emph{{ICLR}}. OpenReview.net.

\bibitem[{Zhou et~al.(2024{\natexlab{a}})Zhou, Xu, Zhu, Zhou, Lo, Sridhar, Cheng, Ou, Bisk, Fried, Alon, and Neubig}]{DBLP:conf/iclr/ZhouX0ZLSCOBF0N24}
Zhou, S.; Xu, F.~F.; Zhu, H.; Zhou, X.; Lo, R.; Sridhar, A.; Cheng, X.; Ou, T.; Bisk, Y.; Fried, D.; Alon, U.; and Neubig, G. 2024{\natexlab{a}}.
\newblock WebArena: {A} Realistic Web Environment for Building Autonomous Agents.
\newblock In \emph{{ICLR}}. OpenReview.net.

\bibitem[{Zhou et~al.(2025)Zhou, Jiang, Tian, Weston, Levine, Sukhbaatar, and Li}]{DBLP:journals/corr/abs-2503-15478}
Zhou, Y.; Jiang, S.; Tian, Y.; Weston, J.; Levine, S.; Sukhbaatar, S.; and Li, X. 2025.
\newblock {SWEET-RL:} Training Multi-Turn {LLM} Agents on Collaborative Reasoning Tasks.
\newblock \emph{CoRR}, abs/2503.15478.

\bibitem[{Zhou et~al.(2024{\natexlab{b}})Zhou, Zanette, Pan, Levine, and Kumar}]{DBLP:conf/icml/ZhouZPLK24}
Zhou, Y.; Zanette, A.; Pan, J.; Levine, S.; and Kumar, A. 2024{\natexlab{b}}.
\newblock ArCHer: Training Language Model Agents via Hierarchical Multi-Turn {RL}.
\newblock In \emph{{ICML}}. OpenReview.net.

\end{thebibliography}

\newpage
\appendix
\section{Training and Evaluation Details}

\subsection{Training Details}\label{sec:app:training_details}
First of all, we perform SFT on Qwen for $2$ epoch, aiming at improving the task-specific instruction following capability and the potential sampling efficiency during the RL stage. Second, we conduct RL training on the finetuned checkpoint for a maximum of $100$ steps. We perform experiments on $4$ A100-40G GPUs with a maximum of $16$k tokens. During the RL training stage, we optimize the policy model using the GRPO algorithm, setting the maximum interaction turns to be $15$ and the maximum length of stepwise environmental feedback to be $2$ due to memory limitation. We remove the KL term to improve training stability, as the gradient explosion phenomenon is observed when perform early experiments. We use VeRL~\citep{DBLP:conf/eurosys/ShengZYWZZPL025} as the main RL training framework. 

\subsection{Training Data Collection Details}\label{sec:app:data_details}
We perform post-training on SWE-bench-extra, a dataset that collect $6,415$ issue-pull request pairs from real Github issues. We filter the dataset based on creation time of the issues, preserve task instance that are created after $2023$, and build up task-specific docker images for each task instance. After construction, we perform unit test through the evaluation method provided by SWE-bench~\citep{DBLP:conf/iclr/JimenezYWYPPN24}, where the ground-truth git patch is applied to the problematic repository, then the unit test is performed to find out whether the issue has been solved. Due to network constraints and machine capacity, we construct $457$ docker images that can successfully pass the test.

To construct the SFT dataset, we utilize \texttt{Deepseek-V3} for trajectory generation from the constructed task instances and their corresponding docker images. For each task instance, we generate $8$ rollouts using the environment and scaffolds we introduced in \S{\ref{sec:environment}} and obtain $3$k trajectories in total. Identical to the model training procedure, we set the number of maximum interaction turns to be $15$ and the number of max generation tokens to be $16$k. We restrict the length of environmental feedback of each interaction turns to be $2$k tokens. During the RL training process, we utilize problems and environments identical to those used to construct the SFT dataset, which is $457$ instances in total.

\subsection{More Details about Evaluation Benchmarks}\label{sec:app:evaluation}
We evaluate the reasoning capabilities of all models in SWE-bench Verified, a human-validated subset of SWE-bench~\citep{DBLP:conf/iclr/JimenezYWYPPN24}, which has $500$ instances curated by OpenAI. We use the environment and scaffolds introduced in \S{\ref{sec:environment}}, and utilize the official docker images provided by SWE-bench to validate the correctness of generated git patches. For all experiments, we report (1) resolution rate (\%), the proportion of resolved task instances, (2) modification rate (\%), the proportion of trajectories where the task repository is modified before calling the Submit scaffold, and (3) completion rate (\%), the proportion of trajectories where the Submit scaffold is called. 

We also perform experiments on kBench~\citep{DBLP:conf/nips/MathaiHMNIYR24}, a datasets that focus on issues about system-level Linux kernel bugs that can result in kernel crash. Specifically, the dataset has $279$ instances, each instance is obtained by Google from the syzbot automatic testing platform. The dataset covers crash issues in several individual subsystems of the Linux kernel to streamline maintenance and development, including network, usb, file system and else. The correct patch content includes various situations such as single-line code modification, single function modification, cross-function modification, and cross-file modification, which conforms to the actual patch in the development process. The execution-based verification requires to recompile the Linux kernel, then re-run the code that can result in kernel crash. We randomly select $50$ cases from the dataset for evaluation~(kBench-50) Similar to SWE-bench Verified, we evaluate models using the environment and scaffolds we introduced in \S{\ref{sec:environment}}.

\subsection{Computational Cost}
We spend about a week for the construction of task-specific docker images. For the trajectory generation procedure, we spend about $3$ days. The SFT process costs about two hours on $4$ A100-40G GPUs, the RL process costs about 24 hours for training $100$ steps. The execution-based verification costs about 12 hours for SWE-bench Verified, and about $33$ hours for kBench-50 in a remote server with 12-core CPU.

\section{More Details about RL Framework}
\subsection{Scaffolds}\label{sec:app:scaffolds}

\paragraph{Shell.} In this work, we relax the restrictions on command types and allow the LM policy to interact with the environment through any Linux command it desired, such as information seeking command \texttt{ls}, packaging installing command \texttt{pip install} and file removal command \texttt{rm -rf}. 

\paragraph{Editor.} The Editor scaffold provide four distinct commands: \texttt{view}, \texttt{insert}, \texttt{replace} and \texttt{create}. The \texttt{view} command opens a window of adjacent lines of code in the specific file, where the window size is decided by specific arguments \texttt{start\_line} and \texttt{end\_line}. The \texttt{insert} command allows LM policy to insert code blocks into a specific location in a file. The \texttt{replace} command allows LM policy to replace current code block by a specific code block. The \texttt{create} command allows LM policy to create a blank file, then initialize the content of the blank file with specific code blocks. 

\paragraph{Web Search.} Once the LM policy generates a human-readable query and sends the query to the Web Search scaffold, the scaffold will return five distinct results, where each result is a long-text summary of the parsed web page with a maximum length of $1,000$ characters. The results are then summarized as an action response and send back to the policy layer. The web search API we utilized in our experiments is provide by LangSearch, and we use the content of field ``summary" in the returned result as the search result.

\subsection{Environment Basics}\label{sec:app:environment}

\paragraph{Image and Packages}
The base docker image we used is \texttt{python:3.11}. Based on the base docker image, we build up task-specific docker images where the python packages are installed in a default miniconda environment ``testbed", which is identical to the settings in the images provided by SWE-bench. The low-level interface layer and the atomic operations are based on the modified version of SWE-ReX~\citep{DBLP:conf/nips/YangJWLYNP24}. Specifically, we extend the waiting duration when creating the deployment, rewrite the failure handling logic to handle timeout errors and update the error message dissemination logic. 

\paragraph{Hindsight Principle} Improper execution of scaffolds and commands may result in task failure. For example, running interactive commands like \texttt{python} and \texttt{vim} with incorrect arguments can lead to session stuck; running potentially dangerous commands like \texttt{rm -rf} can lead to environment collapse, in both cases the subsequent commands will not be successfully executed. We do not set command blacklist to prevent LM policy from executing dangerous commands. In other hand, we follow the hindsight principle, i.e., the early termination of specific task instance caused by task failure and dangerous commands will result in negative rewards with respect to the RL training process.  

\paragraph{Customizable Scaffolds} As introduced in \S{\ref{sec:env:scaffolds}}, the scaffold layer is built upon the low-level interface layer. It is convenient for users to customize specific scaffolds that are compatible with the underlying interface, such as navigation scaffold that incorporating the Linux shell command \texttt{find} and \texttt{grep}. The scaffolds that do not interact with the isolated environment are also compatible, such as remote scaffolds following MCP~(Model Context Protocol). 

\paragraph{Scalable Scenarios} Our framework supports scalable task and scenario instantiation through an automated environment construction pipeline. For each task instance, we dynamically generate a Docker image configured with the necessary dependencies, leveraging the target code repository, a task-specific Git commit ID, and customizable environment settings. To enhance RL compatibility, we decompose the problem-solving process and the outcome verification process, constructing different isolated environments for both process. This separation enables more flexible and scalable reward design while ensuring efficient RL training.

\paragraph{Parallel Generation} Our RL framework supports parallel execution of both the multiple problem-solving processes and multiple outcome verification processes. Specifically, we implement turn-level parallelism during the problem-solving phase: at each turn, the LM policy generates responses for all active instances simultaneously. These responses are then processed by the scaffold layer, converted into executable operations, and executed across isolated environments in parallel. Once all environments return feedback, the context for each instance is updated and passed back to the policy for the next-round response generation.

\section{Template and Cases}\label{sec:app:cases}
\subsection{Chat Template}
\begin{table*}[h!]
    \centering
    \small
    {\ttfamily
    \begin{tabularx}{0.9\textwidth}{X}
    \toprule
    You are a helpful assistant. Interact with the environment via the tools we provide, solve the problem step by step, modifiy the task repository. Finally, call the `submit` tool to submit your changes.\\
    \\
    You are currently at the root of the working directory (i.e., \/testbed). The conda environment (i.e., (testbed)) has been activated.\\
    \\
    Your output format should be as follows:\\
    <think>\\
    <your-thinking-process-here>\\
    </think>\\
    \\
    <tool\_call>\\
    \{"name": <function-name>, "arguments": <args-json-object>\}\\
    </tool\_call> \\
    \bottomrule
    \end{tabularx}
    }
    \caption{The system message for RL training and model inference. The content between "\textless think\textgreater" and "\textless/think\textgreater" is the reasoning part, and the content between "\textless tool\_call\textgreater" and "\textless/tool\_call\textgreater" is the acting part.}
    \label{tab:template:system}
\end{table*}

\begin{table*}[h!]
    \centering
    \small
    {\ttfamily
    \begin{tabularx}{0.9\textwidth}{X}
    \toprule
    \textcolor{blue}{\textless |im\_start|\textgreater system} \\
    You are a helpful assistant. Interact with the environment via the tools we provide, solve the problem step by step, modifiy the task repository. Finally, call the `submit` tool to submit your changes. \textbackslash n\textbackslash nYou are currently at the root of the working directory (i.e., /testbed). The conda environment (i.e., (testbed)) has been activated.\textbackslash n\textbackslash nYour output format should be as follows:\textbackslash n<think>\textbackslash n<your-thinking-process-here>\textbackslash n</think>\textbackslash n<tool\_call>\textbackslash n\{"name": <function-name>, "arguments": <args-json-object>\}\textbackslash n</tool\_call>\textbackslash n\textbackslash nMake sure finish your task within 15 turns, otherwise you will be penalized. Once finished, you should call the `submit` tool to submit your changes.\\
    \textcolor{blue}{\textless|im\_end|\textgreater} \\
    \textcolor{blue}{\textless|im\_start|\textgreater user}\\
    Modeling's `separability\_matrix` does not compute separability correctly for nested CompoundModels\textbackslash nConsider the following model:\textbackslash r\textbackslash n\textbackslash r\textbackslash n```python\textbackslash r\textbackslash nfrom astropy.modeling import models as m\textbackslash r\textbackslash nfrom astropy.modeling.separable import separability\_matrix\textbackslash r\textbackslash n\textbackslash r\textbackslash ncm = m.Linear1D(10) \& m.Linear1D(5)\textbackslash r\textbackslash n```\textbackslash r\textbackslash n\textbackslash r\textbackslash nIt's separability matrix as you might expect is a diagonal:\textbackslash r\textbackslash n\textbackslash r\textbackslash n```python\textbackslash r\textbackslash n>>> separability\_matrix(cm)\textbackslash r\textbackslash narray([[ True, False],\textbackslash r\textbackslash n       [False,  True]])\textbackslash r\textbackslash n```\textbackslash r\textbackslash n\textbackslash r\textbackslash nIf I make the model more complex:\textbackslash r\textbackslash n```python\textbackslash r\textbackslash n>>> separability\_matrix(m.Pix2Sky\_TAN() \& m.Linear1D(10) \& m.Linear1D(5))\textbackslash r\textbackslash narray([[ True,  True, False, False],\textbackslash r\textbackslash n       [ True,  True, False, False],\textbackslash r\textbackslash n       [False, False,  True, False],\textbackslash r\textbackslash n       [False, False, False,  True]])\textbackslash r\textbackslash n```\textbackslash r\textbackslash n\textbackslash r\textbackslash nThe output matrix is again, as expected, the outputs and inputs to the linear models are separable and independent of each other.\textbackslash r\textbackslash n\textbackslash r\textbackslash nIf however, I nest these compound models:\textbackslash r\textbackslash n```python\textbackslash r\textbackslash n>>> separability\_matrix(m.Pix2Sky\_TAN() \& cm)\textbackslash r\textbackslash narray([[ True,  True, False, False],\textbackslash r\textbackslash n       [ True,  True, False, False],\textbackslash r\textbackslash n       [False, False,  True,  True],\textbackslash r\textbackslash n       [False, False,  True,  True]])\textbackslash r\textbackslash n```\textbackslash r\textbackslash nSuddenly the inputs and outputs are no longer separable?\textbackslash r\textbackslash n\textbackslash r\textbackslash nThis feels like a bug to me, but I might be missing something?\\
    \textcolor{blue}{\textless |im\_end|\textgreater} \\
    \textcolor{blue}{\textless |im\_start|\textgreater assistant}\\
    \textcolor{red}{<think>}\textbackslash nThe issue described involves the `separability\_ matrix` function in `astropy.modeling.separable` not correctly computing separability for nested `CompoundModel` instances. The user expects the separability matrix to maintain diagonal or block-diagonal forms for nested models, but it does not.\textbackslash n\textbackslash nTo investigate, I should:\textbackslash n1. Check the source code of `separability\_matrix` to understand how it computes separability for nested models.\textbackslash n2. Verify the expected behavior for nested `CompoundModel` instances.\textbackslash n3. Identify if this is a bug or if there's a misunderstanding in the expected output.\textbackslash n\textbackslash nFirst, I need to locate the `separability\_matrix` function in the astropy codebase. Since I'm working in a testbed environment, I'll first check if the astropy source is available or if I need to install it.\textbackslash n\textcolor{red}{/think>}\textbackslash n\textcolor[HTML]{006400}{<tool\_call>}\textbackslash n\{"name": "shell", "arguments": \{"command": "find / -name \textbackslash"modeling\textbackslash" -type d 2>/dev/null"\}\}\textbackslash n\textcolor[HTML]{006400}{</tool\_call>}\textcolor{blue}{\textless |im\_end|\textgreater} \\
    \textcolor{blue}{\textless |im\_start|\textgreater user}\\
    \textcolor{blue}{\textless tool\_response\textgreater} \\
    Current Turn: 2\\
    \\
    /testbed/astropy/modeling\textbackslash n/testbed/docs/changes/modeling\textbackslash n/testbed/docs/modeling \\
    \textcolor{blue}{\textless /tool\_response\textgreater \textless |im\_end|\textgreater} \\
    \textcolor{blue}{\textless |im\_start|\textgreater assistant}\\
    \textcolor{red}{<think>}\textbackslash nThe astropy modeling source is located at `/testbed/astropy/modeling`. I need to inspect the `separable.py` file within this directory to understand the `separability\_matrix` function's implementation and identify the issue with nested `CompoundModel` instances.\textbackslash n\textcolor{red}{</think>}\textbackslash n\textcolor[HTML]{006400}{<tool\_call>}\textbackslash n\{"name": "editor", "arguments": \{"command": "view", "path": "/testbed/astropy/modeling/separable.py"\}\}\\
    \textcolor[HTML]{006400}{</tool\_call>}\textless |im\_end|\textgreater     \\
    \textcolor{blue}{\textless |im\_start|\textgreater user}\\
    \textcolor{blue}{\textless tool\_response\textgreater} \\
    Current Turn: 3\\
    \\
    Error when using editor and performing command `view`.\textbackslash nReason: arguments `start\_line` and `end\_line` are required. Currently: dict\_keys([`command`, `path`]) \\
    \textless /tool\_response\textgreater \textcolor{blue}{\textless |im\_end|\textgreater} \\
    \bottomrule
    \end{tabularx}
    }
    \caption{The first two turn of interaction between Deepseek-V3 and the environment. The content between "\textcolor{red}{\textless think\textgreater}" and "\textcolor{red}{\textless/think\textgreater}" is the reasoning part, and the content between "\textcolor[HTML]{006400}{\textless tool\_call\textgreater}" and "\textcolor[HTML]{006400}{\textless/tool\_call\textgreater}" is the acting part.}
    \label{tab:template:chat}
\end{table*}

We provide the system message in Table~\ref{tab:template:system}. We provide the interaction history between Deepseek-V3 and the isolated environment for the first two turns for task instance ``astropy\_\_astropy-12907" in Table~\ref{tab:template:chat} due to space limitation. 

\subsection{Cases}
In this section, we provide several failed trajectories that we observed during the RL training and model inference when performing the V-RA method.
\paragraph{Repeated Operation.} The LM policy is likely to generate same operation, or similar operations in multiple consecutive actions. We provide an example in Table~\ref{tab:failure_repeat}. As shown in the example, the LM policy spends many turns of interaction to navigate the problematic file and the problematic code. However, the location of the problematic file has already been mentioned in the issue, and the issue is appended into the input of the LM policy each turn.  
\begin{table*}[h!]
    \centering
    \small
    {\ttfamily
    \begin{tabularx}{0.9\textwidth}{X}
    \toprule
    \textbf{Issue} \\
    \midrule
    Modeling's `separability\_matrix` does not compute separability correctly for nested CompoundModels\\Consider the following model:\\\\```python\\from astropy.modeling import models as m\\from astropy.modeling.separable import separability\_matrix\\\\cm = m.Linear1D(10) \& m.Linear1D(5)\\```\\...\\If however, I nest these compound models:\\```python\\>>> separability\_matrix(m.Pix2Sky\_TAN() \& cm)\\...\\```\\Suddenly the inputs and outputs are no longer separable?\\\\This feels like a bug to me, but I might be missing something? \\
    \midrule
    \textbf{Turn 1: shell-grep} \\
    \midrule
    \{"name":"shell","arguments":\{\\"command":"grep -r "separability\_matrix" | grep -v "astropy.modeling.separable"",\\"expect":""\}\} \\
    \midrule
    \textbf{Turn 2: shell-grep} \\
    \midrule
    \{"name":"shell","arguments":\{\\"command":"grep -r "separability\_matrix" astropy/modeling/separable","expect":""\}\} \\
    \midrule
    \textbf{Turn 3: shell-grep} \\
    \midrule
    \{"name":"shell","arguments":\{\\"command":"grep -r "separability\_matrix" | grep -i "astropy.modeling"","expect":""\}\} \\
    \midrule
    \textbf{Turn 4: editor-view} \\
    \midrule
    \{"name":"editor", "arguments": \{\\"command": "view", "path": "astropy/modeling/separable.py", "newline": "\textbackslash n"\}\} \\
    \midrule
    \textbf{Turn 5: editor-view} \\
    \midrule
    \{"name":"editor","arguments":\{\\"command":"view","path":"astropy/modeling/separable.py",\\"newline":"\textbackslash n","start\_line":"1","end\_line":"100"\}\} \\
    \midrule
    \textbf{Turn 6: editor-view} \\
    \midrule
    \{"name":"editor","arguments":\{\\"command":"view","path":"/testbed/astropy/separable.py",\\"newline":"\textbackslash n","start\_line":"1","end\_line":"100"\}\} \\
    \midrule
    \textbf{Turn 7: editor-view} \\
    \midrule
    \{"name":"editor","arguments":\{\\"command":"view","path":"/testbed/astropy/modeling/core.py",\\"newline":"\textbackslash n","start\_line":"1","end\_line":"100"\}\} \\
    \midrule
    \textbf{Turn 8: editor-view} \\
    \midrule
    \{"name": "editor", "arguments": \{\\"command": "view", "path": "/testbed/astropy/modeling/separable.py", "start\_line": "1", "end\_line": "1000"\}\} \\
    \midrule
    \textbf{Turn 9: editor-view} \\
    \midrule
    \{"name": "editor", "arguments": \{\\"command": "view", "path": "/testbed/astropy/modeling/separable.py", "start\_line": "1", "end\_line": "318"\}\} \\
    \bottomrule
    \end{tabularx}
    }
    \caption{The case that LM policy repeatedly calls the ``grep" command and ``view" command to navigate problematic files~(preserve the first 9 turns only). The LM policy traverses different files in the repository. We simplify the issue description, omit the thinking part and preserve the parsed action for better visualization.}
    \label{tab:failure_repeat}
\end{table*}

\paragraph{Identical Behaviors.} The LM policy can have similar behavior across different task instances, where the actions in the generated trajectories are correct but meaningless. We provides an example in Table~\ref{tab:failure_meaningless}. As shown in the example, the LM policy utilizes commands like ``view", ``ls", `python" and ``echo" to interact with the environment. However, these commands cannot handle the current issue. These ``safe" commands can be easliy requested but also ineffective.

\begin{table*}[h!]
    \centering
    \small
    {\ttfamily
    \begin{tabularx}{0.9\textwidth}{X}
    \toprule
    \textbf{Issue} \\
    \midrule
    collect\_factor\_and\_dimension does not detect equivalent dimensions in addition\textbackslash nCode to reproduce:\\```pythonfrom sympy.physics import units\\from sympy.physics.units.systems.si import SI\\\\v1 = units.Quantity('v1')\\SI.set\_quantity\_dimension(v1, units.velocity)\\SI.set\_quantity\_scale\_factor(v1, 2 * units.meter / units.second)\\\\a1 = units.Quantity('a1')\\SI.set\_quantity\_dimension(a1, units.acceleration)\\SI.set\_quantity\_scale\_factor(a1, -9.8 * units.meter / units.second**2)\\\\t1 = units.Quantity('t1')\\SI.set\_quantity\_dimension(t1, units.time)\\SI.set\_quantity\_scale\_factor(t1, 5 * units.second)\\\\expr1 = a1*t1 + v1\\SI.\_collect\_factor\_and\_dimension(expr1)\\```\\Results in:\\```\\Traceback (most recent call last):\\  File "<stdin>", line 1, in <module>\\  File "C:\textbackslash\textbackslash Python\textbackslash\textbackslash Python310\textbackslash\textbackslash lib\textbackslash\textbackslash site-packages\textbackslash\textbackslash sympy\textbackslash\textbackslash physics\textbackslash\textbackslash units\textbackslash\textbackslash unitsystem.py", line 179, in \_collect\_factor\_and\_dimension\\    raise ValueError(\\ValueError: Dimension of "v1" is Dimension(velocity), but it should be Dimension(acceleration*time)\\```\textbackslash n \\   
    \midrule
    \textbf{Turn 1: editor-view} \\
    \midrule
    \{"name":"editor","arguments":\{"command":"view","path":\\"C:/Python/Python310/lib/site-packages/sympy/physics/units/unitsystem.py",\\"newline":"\textbackslash n"\}\} \\
    \midrule
    \textbf{Turn 2: editor-view} \\
    \midrule
    \{"name":"editor","arguments":\{"command":"view","path":\\"C:/Python/Python310/lib/site-packages/sympy/physics/units/unitsystem.py",\\"start\_line":"1","end\_line":"1000","newline":"\textbackslash n"\}\} \\
    \midrule
    \textbf{Turn 3: shell-echo} \\
    \midrule
    \{"name":"shell","arguments":\{"command":"echo \$PWD","expect":"/testbed","timeout":10\}\} \\
    \midrule
    \textbf{Turn 4: shell-ls} \\
    \midrule
    \{"name":"shell","arguments":\{"command":\\"ls -la /testbed/site-packages/sympy/physics/units","expect":"unitsystem.py","timeout":10\}\} \\
    \midrule
    \textbf{Turn 5: shell-ls} \\
    \midrule
    \{"name":"shell","arguments":\{"command":\\"ls -la /testbed/site-packages","expect":"python3.10","timeout":10\}\} \\
    \midrule
    \textbf{Turn 6: shell-echo} \\
    \midrule
    \{"name":"shell","arguments":\{"command":\\"echo \$PYTHONPATH","expect":"/testbed/python310","timeout":10\}\} \\
    \midrule
    \textbf{Turn 7: shell-pip} \\
    \midrule
    \{"name":"shell","arguments":\{"command":\\"pip list","expect":"sympy==1.10.1","timeout":10\}\} \\
    \midrule
    \textbf{Turn 8: shell-where} \\
    \midrule
    \{"name":"shell","arguments":\{"command":"where pip","timeout":10\}\} \\
    \midrule
    \textbf{Turn 9: shell-ls} \\
    \midrule
    \{"name":"shell","arguments":\{"command":\\"ls -la /testbed/sympy","expect":"unitsystem.py","timeout":10\}\} \\
    \bottomrule
    \end{tabularx}
    }
    \caption{The case that LM policy generates meaningless actions during the interaction with the environment~(preserve the first 9 turns only). We omit the thinking part and only preserve the parsed action for better visualization.}
    \label{tab:failure_meaningless}
\end{table*}

\end{document}